# Introducing Brain-like Concepts to Embodied Hand-crafted Dialog Management System

**Frank Joublin** FRANK.JOUBLIN@HONDA-RI.DE
*Honda-Research Institute Europe*
*Offenbach am Main, Germany*

**Antonello Ceravola** ANTONIO.CERAVOLA@HONDA-RI.DE
*Honda-Research Institute Europe*
*Offenbach am Main, Germany*

**Cristian Sandu** CRISTIAN.SANDU@RINFTECH.COM
*rinf.tech,*
*Bucharest, Romania*

**Abstract**

Along with the development of chatbot, language models and speech technologies, there is a growing possibility and interest of creating systems able to interface with humans seamlessly through natural language or directly via speech. In this paper, we want to demonstrate that placing the research on dialog system in the broader context of embodied intelligence allows to introduce concepts taken from neurobiology and neuropsychology to define behavior architecture that reconcile hand-crafted design and artificial neural network and open the gate to future new learning approaches like imitation or learning by instruction. To do so, this paper presents a neural behavior engine that allows creation of mixed initiative dialog and action generation based on hand-crafted models using a graphical language. A demonstration of the usability of such brain-like inspired architecture together with a graphical dialog model is described through a virtual receptionist application running on a semi-public space.

## Introduction

Over the last decades, a multitude of approaches have been investigated to handle the problem of dialog management in chatroom, in digital assistants and generally in natural language user interface systems. The proposed solutions have been classified in three main categories: hand-crafted models, machine-learning based solutions and hybrid systems. Although machine learning approaches are attractive for their capability to mimic more accurately human dialog naturalness in most cases and to handle a certain degree of openness in dialogs (i.e., chatbot), they suffer from various drawbacks like the amount of dialog data needed to train them, the difficulty to define appropriate reward functions for reinforcement learning solutions or the limited controllability/explainability inherent to statistical-based neural network approaches. Hand-crafted dialogs on the other side, are relatively easy to develop, have a high level of controllability per design and offer good solutions for goal-directed dialogs over well-defined domains. Nevertheless, scalability remains an important issue for all these approaches, either in term of training data or dialog complexity.

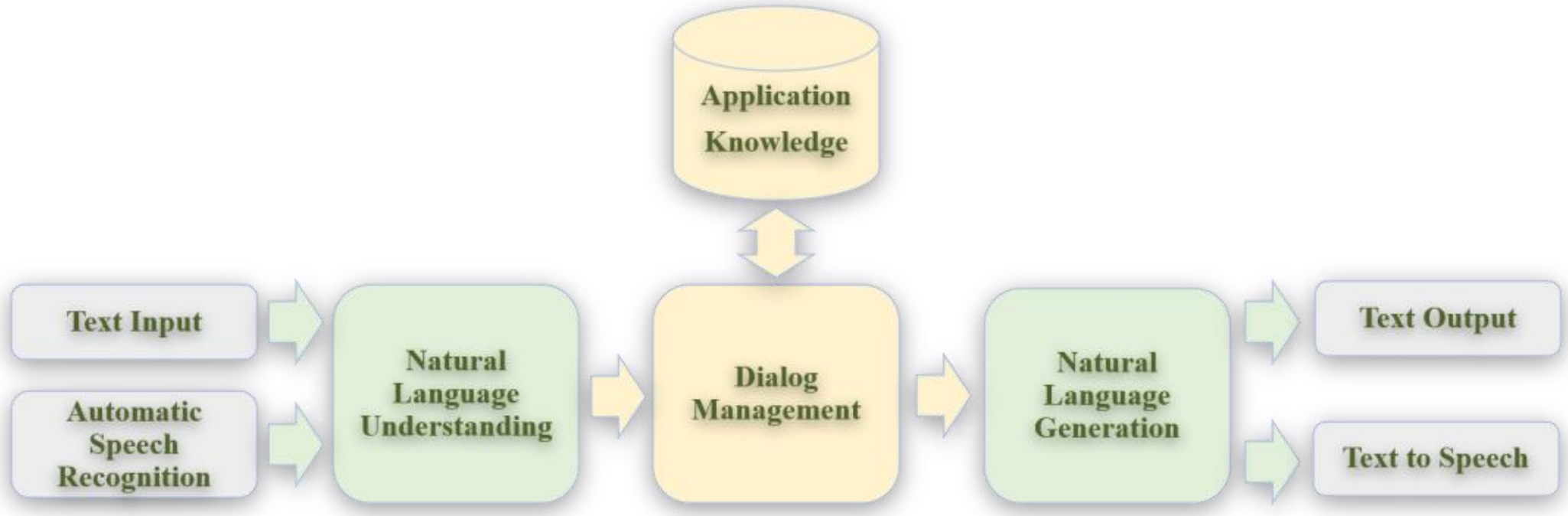

Figure 1: Information flow in traditional speech assistant. The interface of such a system can either be textual input / output like in chatbot or based on speech to text / text to speech modules like in vocal assistant.

Although the usage of such methodologies led to the creation of applications based on natural speech since the mid-60's, mostly in form of chatbot, the introduction on the market of personal assistant like Apple Siri®, Amazon Alexa® or Google Home® (European Commision, 2018) have made speech activated services accessible to everybody. At the same time, applications of such technology are spreading in the industry in a lot of sectors like financial services, insurance (Darda, Chitnis, 2019), e-commerce (Li et Al, 2017), healthcare (Dojchinovski et Al., 2019), air traffic management (Helmke et al. 2015), or automobile (Lugano, 2017), to cite a few. Most commercial products and research architecture are based on a similar computation structure as depicted in Figure 1. It consists of a feedforward flow of information from input to output. Users of such systems can interact via a chat interface in textual form or a directly by voice via speech signal captured from a microphone. In the latter case, the audio signal is analyzed by an automatic speech recognition (ASR) module that transform the signal into a text string delivered regularly as intermediate speech transcriptions or as a whole sentence when the locutor makes a pause. This text string is then processed by a natural language understanding (NLU) module with the role of extracting a representation of the sentence semantic. Such NLU module can be very complex since the semantics of words and phrases might be ambiguous and not only dependent on lexicon and syntax but also on pragmatics, namely the context and mental state of the speaker (see Navigli, 2018, for a critical review of advanced NLU). Nevertheless, most speech assistants to date use simpler NLU (Canonico & De Russis, 2018; Abdellatif, 2020) that mostly extract intents and named entities from sentences. Intents represents a mapping between the user utterance and the inferred intention behind the "speech act" (Morelli et al. 1991). Named entities, instead, are slots of information extracted from the sentence and associated to specific types of information like dates, names, time, place, … The extracted semantic information is provided to a dialog management (DM) module which role is to keep track of the dialog state, context and to decide which speech act should be triggered as a result of the newly acquired information (see Masche & Le, 2018; Burgan, 2017; Lee et al., 2010 for a review of DM systems). In most applications, DM modules needs to access application related data stored in database or knowledge graphs in order to provide pertinent answers to user queries. Finally, the retrieved information is integrated in the generation of textual utterance through a natural language generation (NLG) module. The resulting text is either used as-is in a chat application or sent to a text-to-speech (TTS) system in case of voice enabled applications. NLG modules can be implemented with various degree of complexity (see Santhanam & Shaikh, 2019; Gatt & Krahmer, 2018 for review). The most recent advance in statistical language models using seq2seq transformer models with deep neural network have recently shadowed the traditional grammar-based and template-based realizer of sentence production (Theune, 2003). Nevertheless, the latter are still used in most chat and dialog system of the above cited applications. Their simplicity and real-time capabilities make them suitable for hand-crafted domain specific dialog applications.

Notwithstanding the recent advances in using deep neural architecture for natural language processing, several industrial applications requiring dialog interaction with customers do not have access to huge amount of data or important computational resources to train and run such networks. Our goal in this paper is to show that dialog system can still be constructed based on simple principles, some of them inspired from neuroscience, others from model driven development (MDD) and with low computational requirements.

The first section of this paper will give such an example from neurobiology and try to explain how the concept of mirror neurons (Rizzolatti & Craighero, 2004), can inspire the technological realization of NLU and NLG. The implementation of this concept called "Mirons" in this work, provides designers of dialog systems with a simple way to represent both recognition and production of linguistic intent.

The second section will take a stand on the importance of embodiment (Pezzulo et al., 2011) in social bot either in form of virtual avatars or physical robots and the necessity to define an abstraction layer where multimodal input and output can be treated and represented in a uniform way.

In the third section, the core of the state machine organizing the DM module and more generally the behavior of the embodiment will be described in detail. A description of its architecture, organized around a recursive neural network with long term memory states will clarify its usage.

The next section will illustrate another phenomenon observed subjectively by humans (Alderson-Day & Fernyhough, 2015; Morin, 2012), namely the possibility for the system to speak internally to itself so to trigger procedural reactions in a similar way as if initiated by an outer request.

The following section 5 will describe how the use of MDD modelling (see Rodrigues da Silva 2015 for a review of these concepts) enables dialog designers to easily define the system behavior in a graphical way using a specialized graphical domain specific language (DSL). This language is used to create the model that through code generation defines weights as well as grammatical structure of the Mirons used in NLU and NLG modules.

Section 6 will give some guidelines to structure the design of dialogs.

Then, section 7 will give an example of application using the presented methodology for implementing a virtual receptionist in a public space.

Finally, section 8 will conclude through a discussion on the advantages and limitations of such an approach used to realize social bots and will give a glimpse into possible next steps for scaling dialog design to the next level without compromising simplicity.

## 1. Natural Language Understanding and Mirror Neurons

Mirror Neurons have first been discovered in area F5 of the premotor cortex of macaque (Di Pellegrino, 1992) and later in several other regions like parietal and pre-frontal areas. They represent a category of neurons that get active when the animal makes a particular movement like grasping for an object, for example, but also when the monkey observes a similar movement performed by a conspecific or a human experimenter. The role of such neurons has been at first considered to be responsible for action understanding, imitation, theory of mind (Gallese, 1998) and language development. Recent critics of the theory (Hickok, 2009) are considering mirror systems to be part of larger systems contributing to the above-mentioned behaviors although maybe, not to be causally responsible. Nevertheless, these discoveries also shed light on the evolution of speech in humans:

> "An echo-mirror neuron system exists in humans: When an individual listens to verbal stimuli, there is an automatic activation of his speech-related motor centers." (Rizzolatti & Craighero, 2007).

Although such research seems far from technological implementation of dialog systems, it may provide inspiration for the design of such system. In particular, it brought to our attention the fact that speech understanding, and speech generation could be more interconnected than what is typically exhibited in current systems. Let us now explain how this relation might look like.

### 1.1. Mirror Concept Applied to NLU and NLG

The mirroring concept applied to a dialog can be illustrated as follow: consider someone asking you in the street a question like "Sorry, can you tell me what time it is, please?". Now, the next day imagine yourself in the need of knowing the time and then asking a stranger the exact same question. Could it be that the perception of this question and the production of it share a similar representation? This might or might not be the case in the brain, but this is the assumption we make concerning the technical implementation of it in a dialog-based social bot.

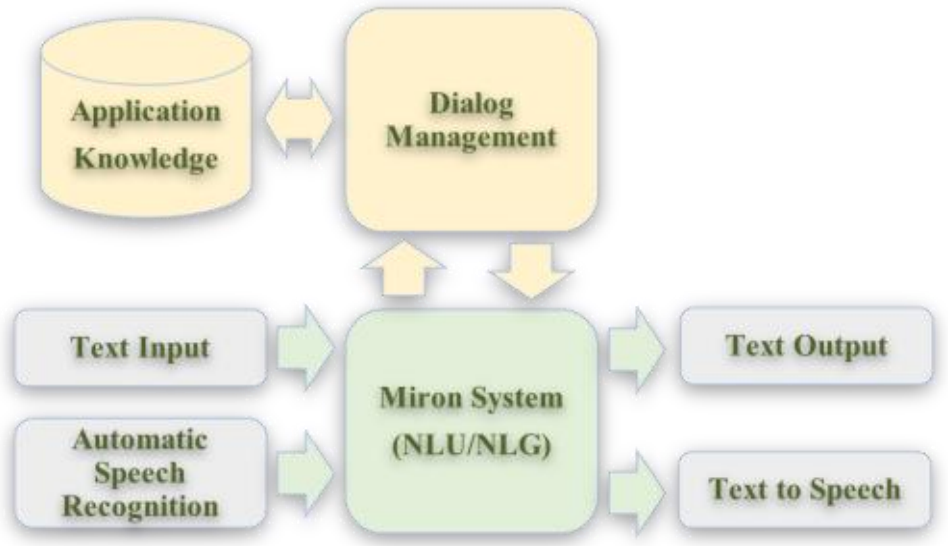

Figure 2: Miron system

As we have seen in introduction, most commercial applications are based on NLU using the concepts of intent and named entities. The former is used to extract the intention behind user utterances and the later to collect pieces of relevant information complementing the intention. In the example:

> "I am looking for a train from Paris to Lyon tomorrow around 12:00."

the intention is to get timetable information about a train, while the extracted named entities could be the departure station (Paris), the destination station (Lyon), the date (tomorrow) and the time range (12:00).

In the case of NLG of the same utterance, most domain specific systems rely on template generation because of their simplicity and their real-time capabilities. In this particular case, the production template could look like:

> "I am looking for a train from **$departure** to **$destination $day** around **$time**."

As can be seen in these two simple examples, the extraction of the information necessary to analyze the utterance (first example) can be identical to the information necessary to produce this same utterance (second example). This fact has been triggering the development of an architecture inspired by the mirror system to relate NLU and NLG together. This allows designers to model once the utterances for both systems at the same time. The next section will describe this architecture more precisely.

## 1.2. The "Miron" System

We called "Miron", a contraction of "Mirror Neuron", a data structure defining a particular intention (or intent) associated with a list of template sentences used both to perceive and to generate this intention in a dialog system. Moreover, Miron definitions are associated to an optional data structure that contains so called "slots" which characterize the named entities used by the template sentences. Such slots represent the information to be extracted from perceived utterances or are used to generate specific to fill specific information in generated utterances. In addition, Mirons mas be associated with "data slots", a concept that allows storing constant information in form of key/value pairs. Data slots can be used to characterize the information retrieved by a Miron or to modify states in the dialog. We will explain this point in more detail in section 3.2.2. To ease the design of dialogs through the use of Mirons, we introduced a DSL with an intuitive syntax (i.e., Behavior DSL) to define the sentence templates and slots associated with an intent. An example of a Miron template written with this DSL can be found in Figure 3. In this example, the intent of the Miron is to "request a train connection" and it is associated to a single template (although several templates could be defined). The template is defined by a readable sentence punctuated with different parenthesis used to annotate the structure of the sentence. Parts of the template embraced with "( ) " are optional. This means, in case of perception, that this Miron can be recognized when the full template matches or if a fraction of it is used as utterance (e.g., "I am looking for a train to Lyon"). In a mirroring fashion, the same Miron may generate the full template or only fractions of it. The definition of optional parts can be nested, giving an easy way to model in a single template a large set of combination of sentences (i.e., allowing generalization). Parts of the template embraced with "< >", called "grammar fields", allow the definition of hierarchical grammatical structures of alternative formulations. The combinatoric created by grammar fields significantly increases the coverage of formulations that a Miron can recognize and generate. Named entities or slots are de-fined by parts embraced with "{ }". A key/value pair creates the mapping between the part of template and the name of the slot in the system (i.e "Paris": part of template→ "Departure": name of the slot).

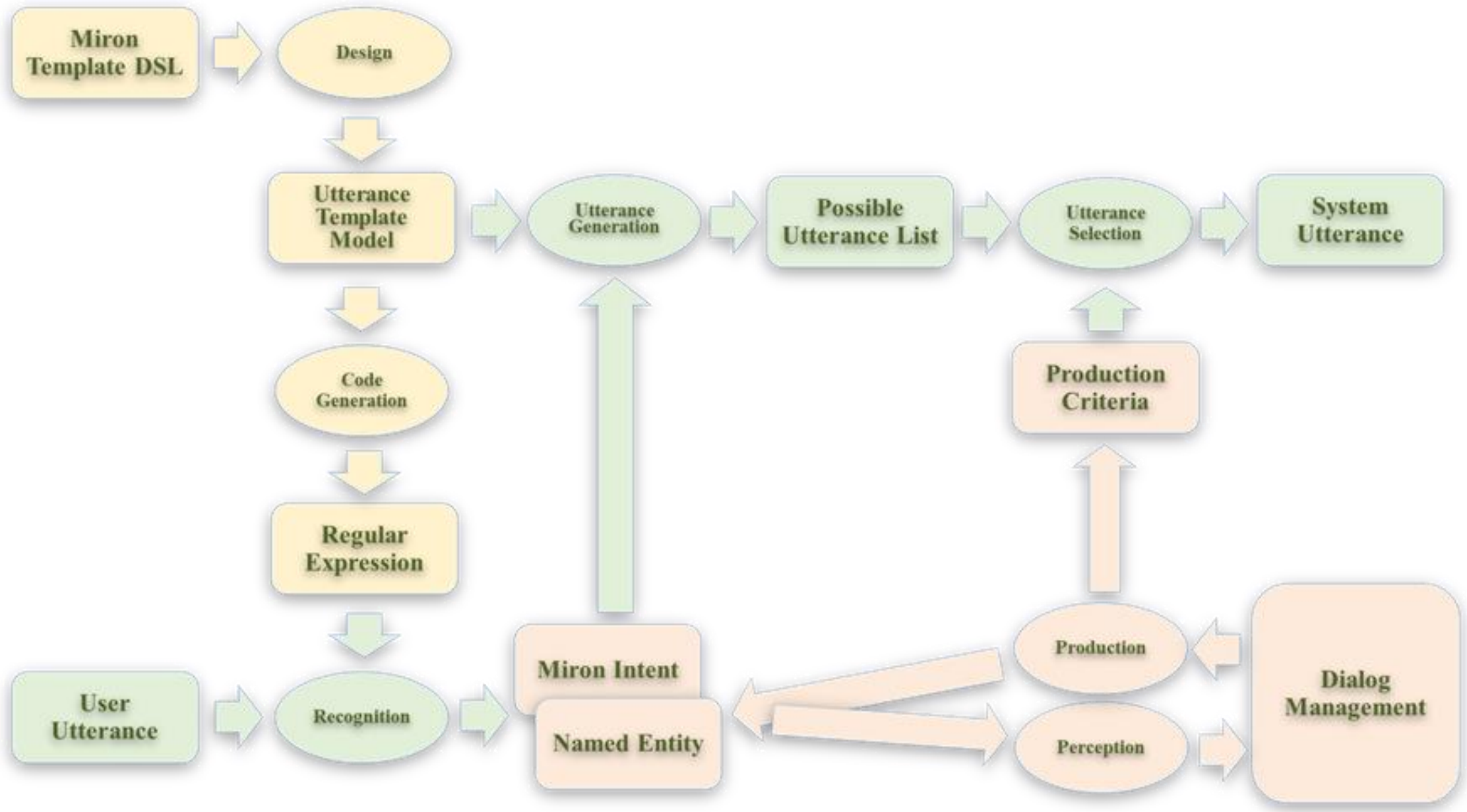

Figure 3: Architecture of the Miron system

The architecture of the Miron system is depicted in Figure 3. The part in yellow in the figure describes the process used at design-time to define Mirons. Utterance template model (like the example of Figure 3) are first created using the above explained Behavior DSL. All the Miron templates are then translated (by a code generation phase) into regular expressions used in the recognition phase and for the extraction of intent and named entities at run-time. The same template model of a given Miron intent together with a set of slots associated to their values is used to produce all the possible utterances defined by the combination of the grammar field and optional parts. Incomplete utterances using undefined or empty slots are suppressed. One among all possible combinations of the template will be selected to produce the system utterance. The selection is most of the time done at random. However other production criteria can be used, such as based on probabilities of selection, choice among different foreign languages, style of language, or the emotional expressiveness to be used. Of course, the criteria for the selection of utterance should be related to information present in the data structure of all Miron. The interface to the dialog management depicted in the reddish part of Figure 3 is restricted to the Miron intent and the slots either as input in case of recognition, or as output in case of production. In the latter case, a production criterion can be provided by the DM to influence the production.

### 1.3. Extension to Statistical Method of the NLU Module

Using regular expressions to analyze user utterance is a relatively simple approach to perform this task, however still used in most commercial (Nuance, IBM) and open-source framework (RASA, spaCy, NLP.js), particularly for the extraction of named entities. To increase generalization, we used the generative property of the Miron templates (i.e., the capacity to generate a set of sentences from each template) to train a language model used to detect intent. In our case, we used NLP.js but the principle is mainly the same with other libraries. This process is depicted in Figure 4.

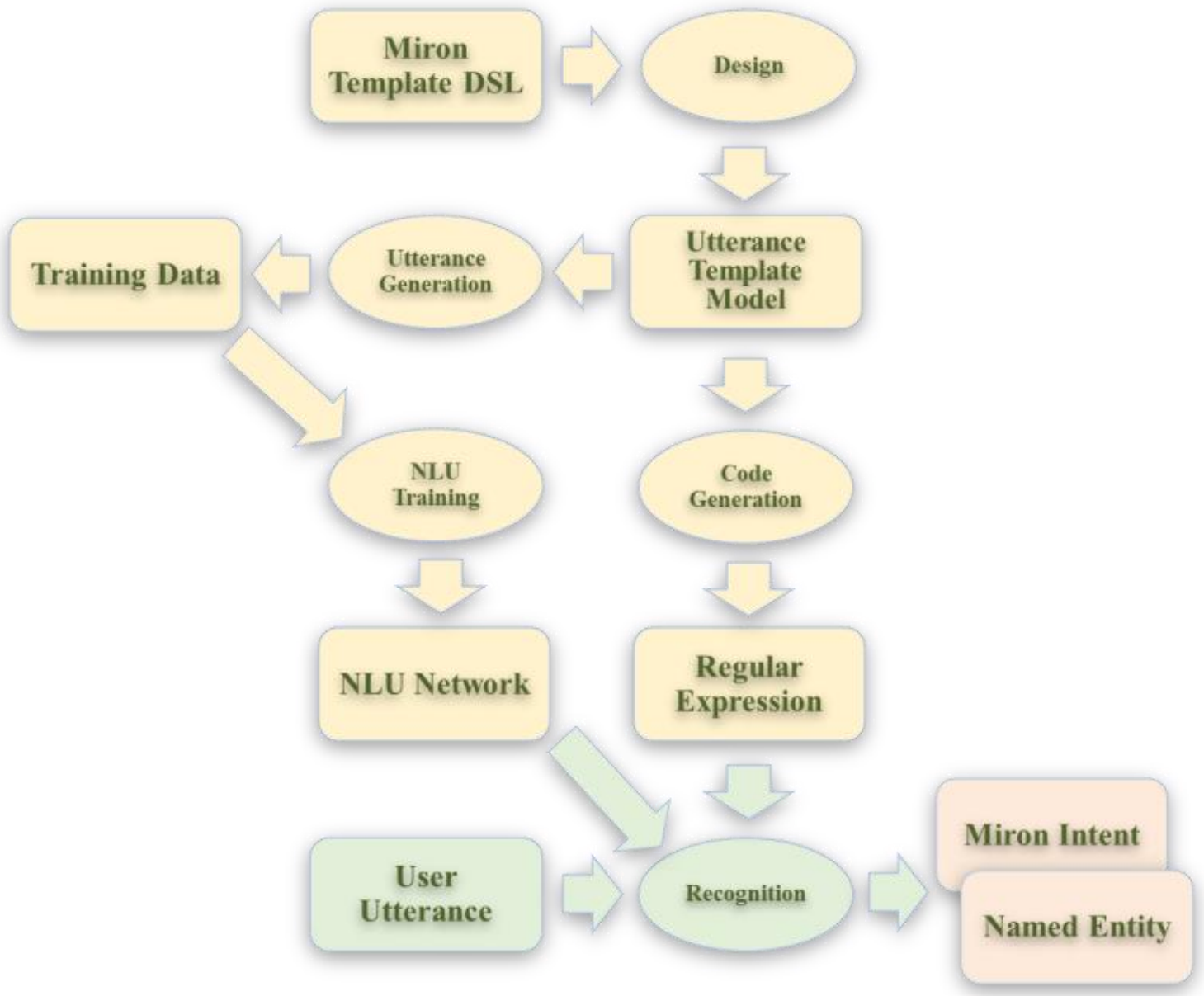

Figure 4: Using state of art NLU in the Miron system. At design-time, the utterance template model of Mirons can be used to generate training data for a statistical approach of NLU. The trained parser can then be used to increase the generalization capabilities of the Miron intent recognition.

The advantage of this approach is that no extra effort is requested from the designer to improve recognition since the training data can be generated from the combinatoric of the grammar fields and the optionality used in each Miron templates.

### 1.4. Advantages of the "Miron" concept

As previously described, the Miron concept is used for utterance definition in a dialog system, where a sentence can be defined once and used in both perception (system input) and generation (system output). This characteristic is particularly relevant in dialog systems that exhibit the following properties:

- Dialogs in which several utterances are shared between machine and user, like for example: greetings, excuses, opinions… In such case, the creation of rich templates via the defined Behavior DSL, allows to exploit expressive variability in both input and output which simplify the design by unifying both interfaces.
- Dialogs where a user reaction model is built to predict user utterances compatible with a system question, for example. Such a user model would require a representation of user utterances in the machine.
- Dialogs where per construction the role of the machine and the user can be switched (e.g., games, teacher/model…)
- Dialog where the generative power of the Miron templates, could be used to automatically generate training samples for statistical NLU models with minimal design effort (as described in §1.3).

Another characteristic of the Miron system, which was driving its conception, is the possibility to use the perceived Mirons to learn dialog behaviors through observation and imitation. This possibility has great potential for future research on learning human machine interaction by demonstrations. However, this is a domain out of the scope of the present paper.

In the next section we briefly discuss the possibility to extend the concept of Mirons to embodied application and to multi-modal interactions.

## 2. Embodiment & Multi-modality

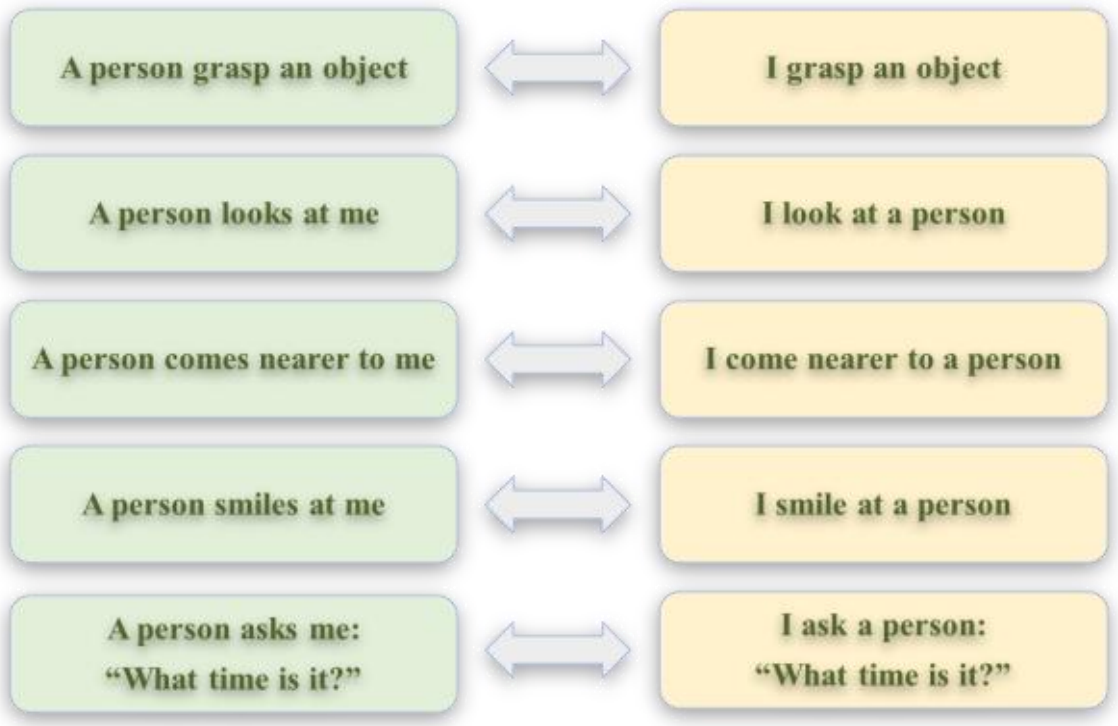

Figure 5: Extension of the Miron concept to other modalities

Chatbot applications of dialog system are often restricted to verbal interactions and sometimes display buttons to propose specific choices for well-defined inputs (Chaves & Gerosa, 2021). This has the advantage of reducing understanding errors while being a well-established way of human-machine communication. Nevertheless, research prototypes and commercial applications also use the possibility to embody the AI agent, in the sense defined by (Deng et al., 2019) in form of virtual avatar (Noh & Hong, 2021) or physical robot to improve user experience (Powers at al., 2017). In such a case, the interfaces must not only cover verbal input/output but also allow the control of other modalities like gaze, facial expression, gestures, …

To control multiple modalities in an embodied avatar or even a robot, it makes sense to define an abstraction level that can interface high-level and low-level processes. Using symbolic textual representation is a simple way to achieve this target. Recent research has shown how textual description of images (Bernardi et al., 2016) or videos (Yao et al., 2015) could be implemented using deep learning approaches. Although still research in progress, these techniques let foresee, that textual representations are getting momentum.

The Miron system defined in the previous sections, can be used to handle multimodal interactions when those are represented in a symbolic, textual form (Figure 5).

The next section will now dive deeper into the core architecture used to organize dialog and behaviors.

## 3. Dialog/Behavior Engine

When designing dialog system, an architecture able to handle asynchronous perceptual events (i.e., speech utterances, sensory signal about the current situation) and produce appropriate verbal or physical reactions becomes essential. Whatever the used method, at the end, the system fundamentally represents a state machine where the future state of the system is a complex function of perceived and own states. State machine can be implemented in several ways. Recurrent Neural networks (RNN) have been shown to be able to represent finite state machine (FSM) (Das 1994, Gile et al. 1995). Although such works are focused on the learning of FSMs by RNNs and the restrictions it poses, we choose to investigate the creation of hand-crafted FSM on a RNN architecture (a behavior engine) modified in such a way to allow concurrent parallel states and non-deterministic decisions.

### 3.1. Recurrent Neural State Machine Concept

Traditional RNN are generally composed of three layers: an input layer, a hidden layer or state layer and an output layer. Although deep architecture can extend the hidden layers to multiple layers, we focus here on a simple three layers model (see Figure 6, left). Our states are defined in terms of rules. Each rule is a state that can be active when its condition becomes valid, at which point a set of associated actions can be performed. The input layer corresponds to the conditions that can trigger specific states or rules. The output layer corresponds to the actions of the active rules (see Figure 6, right). The state layer is reinjected on the next cycle to allow rules to be chained (i.e., to have their condition dependent on previous states). Since we want to be able to hand-craft rules, we must define them in an easy understandable way for human. One of the criteria we used for achieving that is the usage of first order logic operator like AND and OR in rule conditions. In the RNN architecture this is done by decomposing the rule layer in three steps s (see Figure 7): an AND layer, a preselection layer and an OR layer. The AND and OR layer are very similar to sigma-pi network (Plate 2000) from the functional aspect although implemented in a different way. The preselection layer was introduced to allow non-deterministic decisions. The condition layer is composed of four types of information: the events related to recognized Miron intents $\boldsymbol{c}_{i,1}^{MI}(t)$, the state of named entity $\boldsymbol{c}_{j,1}^{NE}(t)$, the feedback events of action completion $\boldsymbol{c}_{r,1}^{AF}(t)$, and the working memory states $\boldsymbol{c}_{l,1}^{WM}(t)$. The concatenation of all these inputs forms the input layer $\boldsymbol{c}_{n,1}(t)$ of the RNN. A weight matrix $\boldsymbol{w}_{k,n}^{COND}$ specifies how this input contributes to the rule activation $\boldsymbol{r}_{k,1}^{AND}(t)$ in combination with the currently active rules $\boldsymbol{r}_{m,1}(t)$ through the weights $\boldsymbol{w}_{k,m}^{RULE}$ :

$$\boldsymbol{r}_{k,1}^{AND}(t) = f\left(\boldsymbol{w}_{k,n}^{COND} \cdot \boldsymbol{c}_{n,1}(t) + \boldsymbol{w}_{k,m}^{RULE} \cdot \boldsymbol{r}_{m,1}(t)\right)$$

where $f(\boldsymbol{x})$ is a step function that operates elementwise and is defined by $f(x) = \rho(x - \theta), \theta = 1 - \varepsilon$, $\varepsilon$ is a small positive real number (i.e., 0.001) used to account for floating point summation approximations and

$$\rho(x) = \begin{cases} 0, & x \le 0 \\ 1, & x > 0 \end{cases}$$

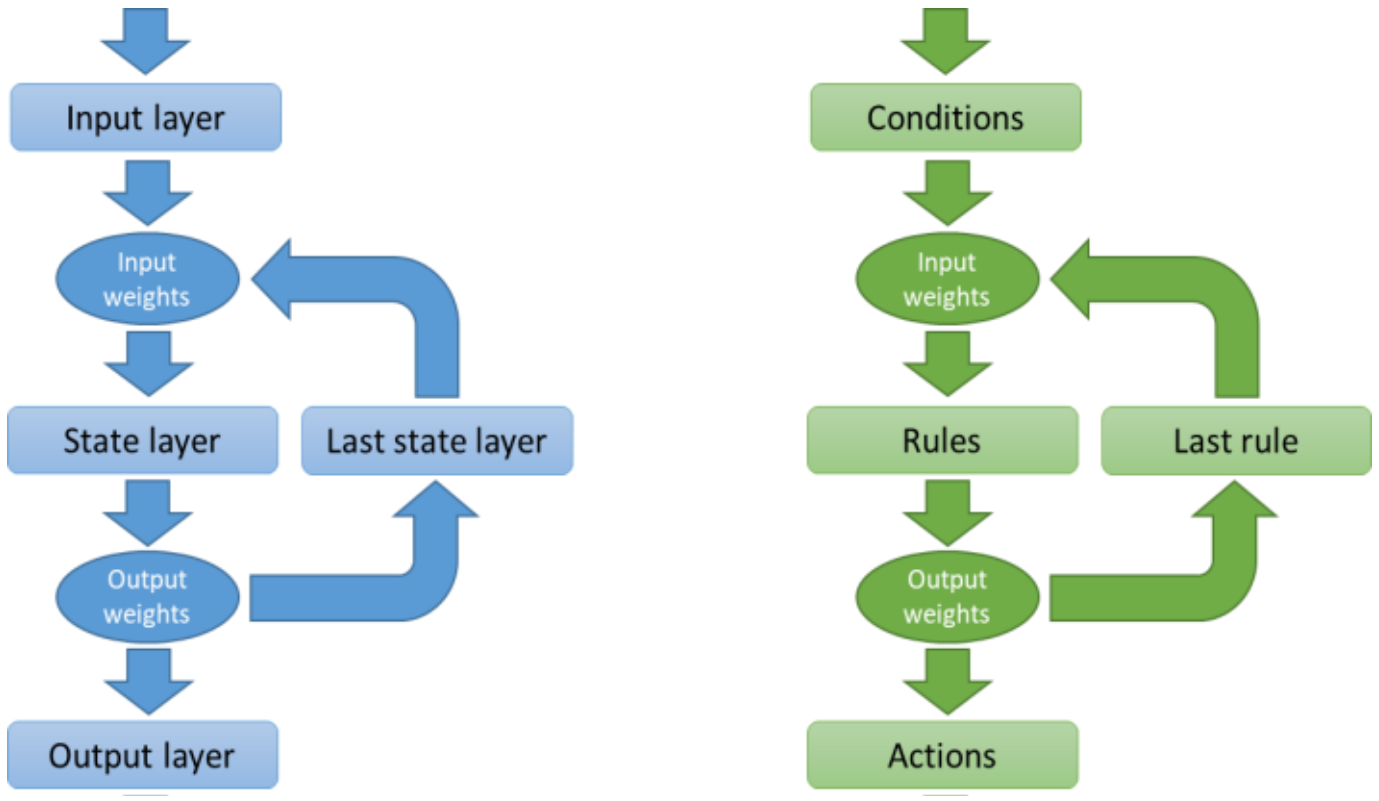

Figure 6: Simple RNN architecture (in blue) and it’s mapping to a finite state machine defining a rule system (in green)

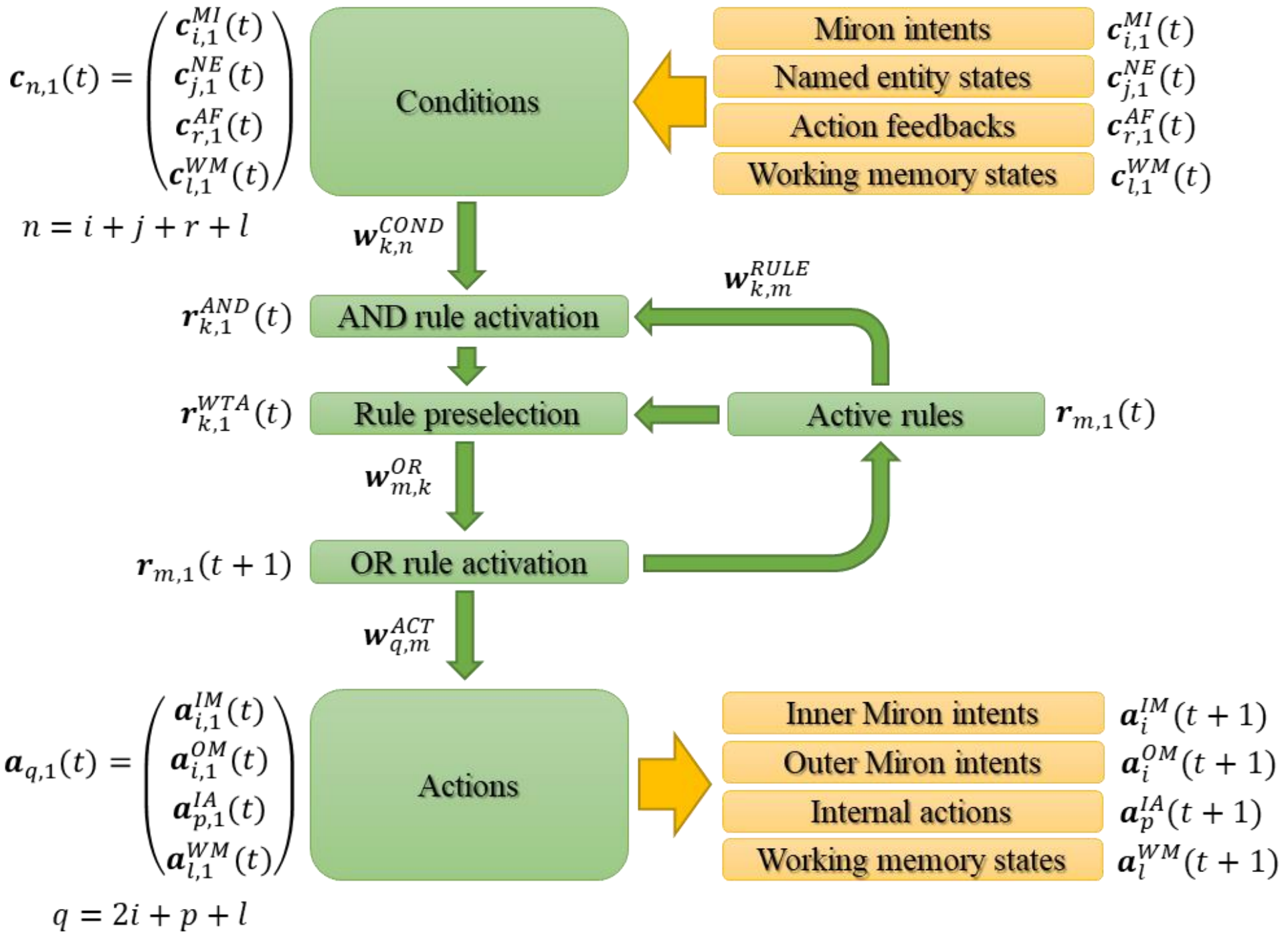

Figure 7: Model of the RNN used (see text for explanation). Notation: vector and matrices are in bold with their dimensions given as subscripts and eventually a name written in superscript.

The weights of an AND condition are defined in such a way that their sum is equal to 1 so that the AND cell activity is above $\theta$. Each active rule which contributes to activate the AND part of multiple next rules selects randomly which of the next rule should be activated. This principle is described by the following computation:

$$\boldsymbol{r}_{k,1}^{WTA}(t) = f\left(\sum_{m} \underset{k}{WTA}\left(\rho(\boldsymbol{w}_{k,m}^{RULE}) \odot (\mathbb{1}_{k,1} \cdot \boldsymbol{r}_{m,1}(t)^{T}) \odot (\boldsymbol{r}_{k,1}^{AND}(t) \cdot \mathbb{1}_{1,m}) \odot (\mathbb{1}_{k,m} + \boldsymbol{\eta}_{k,m}(t))\right) + \boldsymbol{w}_{k,n}^{COND} . \boldsymbol{c}_{n,1}(t)\right)$$

Where $\cdot$ is the matrix multiplication operator, $\odot$ is the element-wise multiplication, $\mathbb{1}_{k,m}$ is a all-ones matrix of dimension *k,m*, $\boldsymbol{\eta}_{k,m}(t)$ is a matrix of small random values changed at each iteration, $\sum_m \boldsymbol{x}_{k,m}$ is the sum over the second dimension of matrix $\boldsymbol{x}$, and $WTA_k\left(\boldsymbol{x}_{k,m}\right)$ is an operator that performs a winner-takes-all along the first dimension of $\boldsymbol{x}$:

$$\underset{k}{WTA}\,(\boldsymbol{x}) = \delta\left(\boldsymbol{x},\ \mathbb{1}_{k,1} \cdot \max_{k}(\boldsymbol{x})\right)$$

with $\delta$ the Kronecker delta.

This selection process allows the system to choose at random among equally possible next states. Once the rule preselection has been caried out, the final new state is determined not only considering the OR part of the conditions but also by the inhibition of rules defined by negative weights in $\boldsymbol{w}_{m,k}^{OR}$:

$$\boldsymbol{r}_{m,1}(t+1) = f\left(\sigma\left(\boldsymbol{w}_{m,k}^{OR} \cdot \boldsymbol{r}_{k,1}^{WTA}(t)\right)\right) \odot \left(\mathbb{1}_{m,1} - \sigma\left(-\boldsymbol{w}_{m,k}^{OR} \cdot \boldsymbol{r}_{k,1}^{WTA}(t)\right)\right)$$

where $\sigma(x) = \rho(x) \cdot x$ is a ramp function. The weights of an OR part of the rule are defined to be either 1 or -1. Each entry associated with a weight of 1 is sure to reach the threshold $\theta$. Negative weights always take the lead by inhibiting the cell.

From the new rule state, the network computes the action $\boldsymbol{a}_{q,1}(t+1)$ to perform using the fan-out matrix of weights $\boldsymbol{w}_{q,m}^{ACT}$:

$$\boldsymbol{a}_{q,1}(t+1) = f\left(\boldsymbol{w}_{q,m}^{ACT} \cdot \boldsymbol{r}_{m,1}(t+1)\right)$$

The action vector is providing activations for four types of actions: inner $\boldsymbol{a}_{i}^{IM}(t+1)$ and outer $\boldsymbol{a}_{i}^{OM}(t+1)$ Miron intents, internal actions $\boldsymbol{a}_{p}^{IA}(t+1)$, and changes in working memory states, $\boldsymbol{a}_{l}^{WM}(t+1)$. We will come back in the next paragraph about their role. Designing behavior or dialog interaction, in this context, means defining the weight matrices $\boldsymbol{w}_{k,n}^{COND}$, $\boldsymbol{w}_{k,m}^{RULE}$,and $\boldsymbol{w}_{q,m}^{ACT}$. Although, these matrices could be learned, we keep this process for future investigations and focus in the present paper on the way to hand-craft these matrices in an easy way (see chapter 6).

### 3.2. Architecture

Figure 8 depicts the interplay between the Miron system and the behavior engine as well as the major components used to define dialog behaviors. Let us review the role of each component:

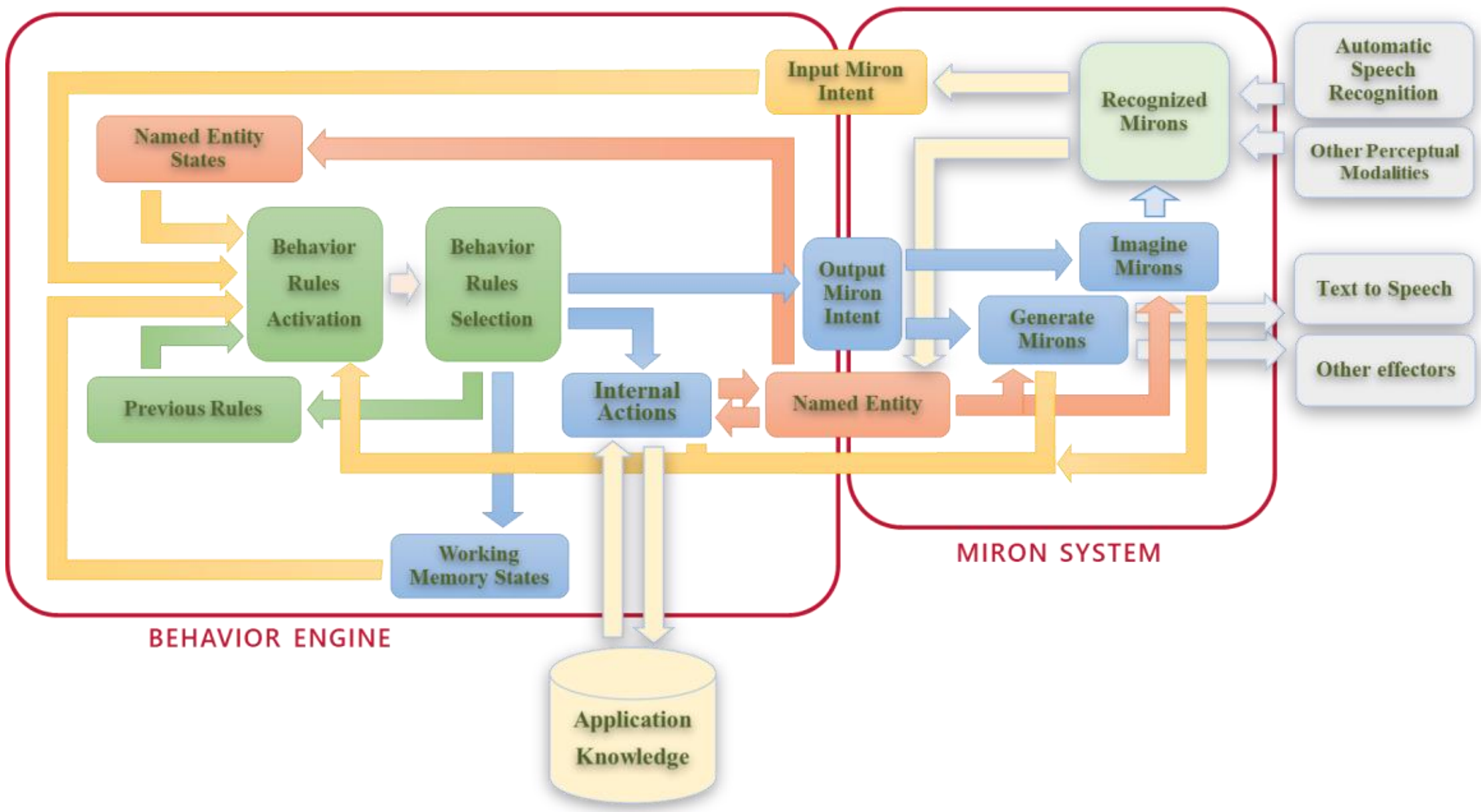

Figure 8: Architecture of the behavior engine

### 3.2.1. Input / Output Mirons

Miron intents (orange and blue blocks in the center of Figure 8) recognized by the Miron system are represented by activations within the condition vector described before. Activations takes the value 0 or 1 depending on the presence of the corresponding Miron. They correspond to recognized Miron templates from external modalities but also from inner Mirons generated by the system itself as described in part 4.

### 3.2.2. Named Entities (slots)

Named entities (red block in the center of Figure 8) are either extracted from external modalities during the Miron recognition process or created/modified by internal actions for example as a result database queries. Rule conditions can be built from the state of named entities (red block on the left of Figure 8). These states represent activations which takes value 0 or 1 if a named entity changed its state, becoming filled or empty, or if its value has changed or not in the last iteration. This means that for each named entity, 4 state variables are created (e.g., filled, empty, changed, unchanged).

### 3.2.3. Rule Activation, Selection and Action Generation

These modules (in green in Figure 8) which activate rule states from the different input conditions (yellow arrows in Figure 8) and determine the actions to be executed in the next cycle (blue arrows in Figure 8) have been extensively described in §3.1.

### 3.2.4. Working Memory States

Working memory states are created to memorize specific events or represent contextual states. They are defined by the level of activity of two type of variables per working memory states: activated memories and inhibited memories. Each of them taking a value of 0 or 1. The combination of the two excludes to have both at level one but does allow to have both at level 0 which correspond to a reset state. It is important to note that this reset state cannot be tested in a condition since resetting a memory state corresponds to removing the state from memory.

#### 3.2.5. Internal Actions

Internal actions are represented as a special type of Mirons that can trigger the execution of predefined functions. These functions can be used for example to access databases or external services, read or write named entity, or any internal functionality required by the application.

In the following section we intent to discuss some aspects that inspired the structure of the behavior engine and some of its properties.

## 4. Action Imagination

One fundamental aspect of human thinking processes is the possibility to follow internally a succession of thoughts, either to recollect memories or imagine present or future scene. Recent discoveries (Mullally & Maguire, 2014) have shown that all these activities might be related to the hippocampus a structure involved in episodic memory and spatial localization and its connection to the core system: a set of brain areas involved in resting state, representation of the self as well as the distinction between real and fictitious experiences (Hassabis, 2007). In this part, we want to get inspiration from some principles, like multiple feedback loops, to give DM the possibility to use, for example, such loops to emulate inner speech.

### 4.1. The role of feedback loops

The multiple roles of feedback loops in the brain have been studied for decades and their role extend from amplification, sustaining activity, action control to attentional process and consciousness (Mashour, 2019).

The central part of the behavior engine (in green in Figure 8 and in Figure 7) is involved in at least five different feedback loops:

- At the deepest level, the feedback loop described in §3.1, allows the system to follow contextual sequences, where the new states of the system depend on the previous ones.
- The working memory state loop described in §3.2.4, allows the system to define and keep long lasting memories of specific conditions or events reflecting dialog states, or contextual memories.
- Long term memory feedback loops can be implemented through the modification of named entity (variables) by internal actions. For example, a rule triggering an internal action making a database queries (application knowledge in Figure 8) could as a result of the query modify the variable that in turn would be part of the condition of other rule activations.
- The feedback loop from the Miron generation process indicating action completion is fundamental to implement asynchronous sequence of behaviors where the next actions depend on the termination of previous actions or internal “thoughts”.
- The most external loop is the perception/action feedback loop that makes dialogs and interaction with the environment possible.

We introduced a sixth loop inspired by the human phonological loop that will be explained now in the next paragraph.

### 4.2. Inner Speech

The concept of "inner speech" is introduced by Vygotsky (Vygotsky, 1986) to characterize speech development in infant from "external/social" to "egocentric/private" to "inner speech". "Private speech" refers to speech that is not produced for the benefit of anyone other than the speaker. This Russian psychologist studied the self-regulatory functions of "inner speech" including altering one's behavior, resisting temptation, regulating mood, making choices, filtering information but also setting goals, planning, solving problems and motivating oneself (Morin, 2013).

Here, we introduce "inner speech" as a way to trigger system behavior in a similar way, independently if the speech signal comes through sensors or as an internal intent from and to the system itself. Recalling the example of §1.1 if in a particular interaction scenario, the system needs to know the time it can ask itself as if it would have been asked externally "What time is it?" and use the result of its internal query to solve the problem at hand. The concept is not fundamentally different from calling an internal function; however, it makes internal processes transparent and easily understandable to anybody inspecting internal states/logs of the system and moreover it makes no difference when an actual user of the system would ask the same question.

We have described until now the major concepts and principles defining the internal functionality of the system, in the next section we are going to describe the methodology we used to create dialogs/behaviors and applications.

## 5. Model Driven Development for Behavior Generation

Designing dialog or behavior applications using the above-described behavior engine requires to define the following information:

- The set of Mirons used to recognize or produce sentences including intent and named entity (slots and data slots)
- The set of weights (as described in §0) encoding the rules defining the logic of the multi-modal behavior/dialog

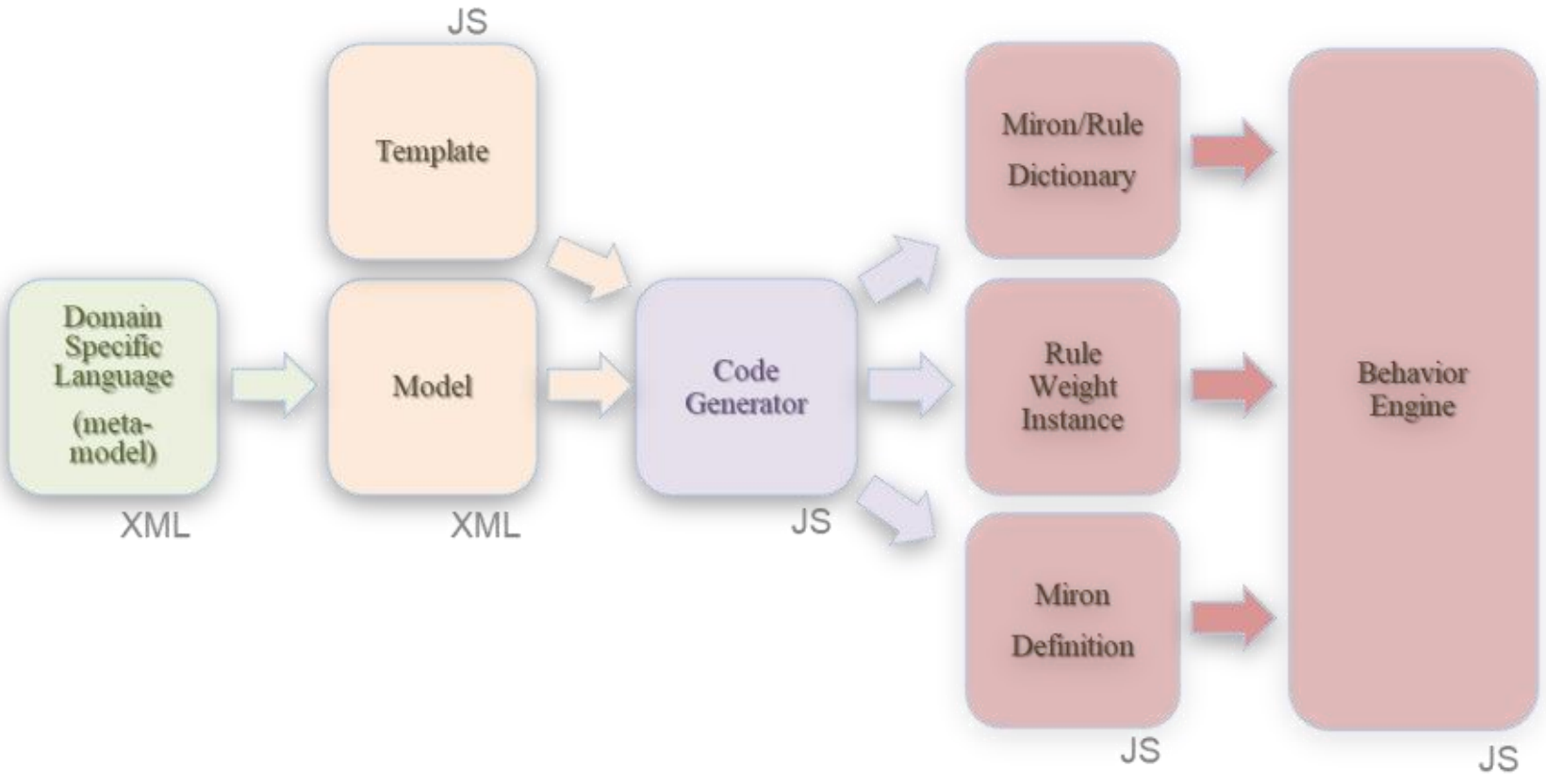

Figure 9: Modeling process used to generate dialogs and avatar behaviors

Such information is represented in the system via a set of three data structures: Miron/Rule Dictionary, Rule Weight Instance and Miron Definition. To simplify this process which would be extremely error-prone and complex to do by hand, we created a specific graphical DSL (meta-model). This language allows the designer to focus on the behaviors and on the logic of the dialog by creating a graphical representation (model).

Figure 9 describes this process: starting from a graphical DSL (described in the next paragraph) a developer creates a model of the dialog/behavior of an application in form of a graph, then a code generator (based on a set of templates) parses the XML file of the graphical model and generates 3 files required by the behavior engine for executing the application:

- A Miron definition file: with all information associated to Miron
- A Rule definition file: containing rule weights and connection definitions for the behavior engine network
- A dictionary file: creating bidirectional mapping tables between Miron/Rule names and unique indexes used for the representation of the behavior engine network

All these JavaScript files are used at initialization of the behavior engine to configure the network.

## 5.1. Graphical Language

The graphical language used to create the behavior DSL is composed of two parts:

- A Miron set of DSL elements allow to define all language related aspects (intent, named entity or slot and Miron templates)
- A Rule set of DSL elements allow to create behavioral logic based on perception conditions and an associated set of actions to be triggered

These two sets of DSL elements intersect at the level of Mirons since Mirons define all perceptions and all actions.

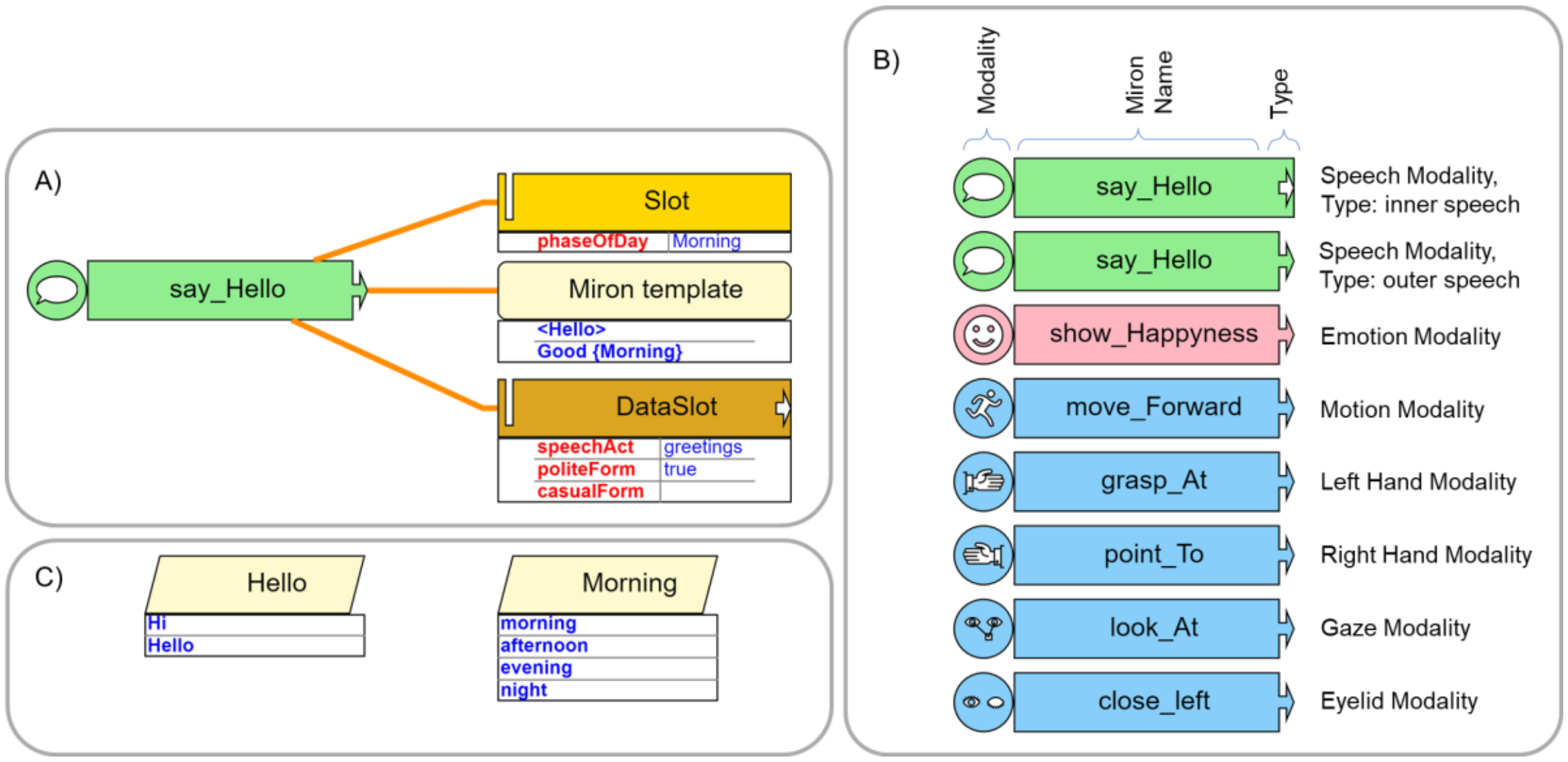

Figure 10: Miron DSL elements: A) Graphical representation of a Miron defined by a modality, a name, a type (inner or outer), templates, named entities (slots) and associated data (data slots). B) Example of different Miron modalities. C) Grammar fields.

### 5.1.1. Miron DSL elements

The graphical representation of Mirons is shown in Figure 10. The green box in part A of the figure represents, for example, a Miron named "say_Hello" associated with the outer speech modality, whose templates covers the following possible phrases: "Hi", "Hello", "Good morning", "Good afternoon", "Good evening", and "Good night". The pattern "{Morning}" corresponds to a name entity that can be extracted from input speech to fill the variable "phaseOfDay" or can use the content of this variable to generate one of the previous sentences. Finally, some data are associated to this Miron in form of three variables: "speechAct" filled with "greetings", "politeForm" filled with "true", and "casualForm" filled with "false". The Miron can be connected to the template, slot and data although all of these objects are optional. In case a Miron has no template defined, the template is set by default as the Miron name.

By defining graphs using this DSL notations the complete grammar of a system can be designed.

### 5.1.2. Rule DSL Elements

Rules, as explained in §0, are triggered by conditions and can trigger actions when the corresponding rule gets activated. The example given in part A of Figure 11 depicts a simple rule showing its graphical representation. In this example the rule can be read like this: if the outer Miron "say_Hello" is perceived (from chat or ASR input) AND the variable "politeForm" is defined and filled (not empty) AND the internal state "greetingsExpected" is true THEN set the internal state "greetingsExpected" to false AND generate the Miron "say_Hello" (to chat or TTS). In this example, since the rule 107 is not connected to any other rules, once activated the rule will be deactivated at the next time step.

Another important aspect of the rule DSL elements concerns the usage of logical operations and chaining of rules. Let us take the example on the left side of Figure 12. In this example, rule 125 is similar to rule 107 of the previous figure, but its condition has been extended to give an example of more complex conditions. All incoming links that connect directly to a rule are managed like a logical OR. So, the condition of rule 125 can be read like this: the rule is activated if at least one of the following conditions is true:

- the outer Miron "say_Hello" is perceived (from chat or ASR input) AND the variable "politeForm" is filled AND the internal state "greetingsExpected" is true OR
- an inner speech Miron "say_Hello" is perceived OR

- rule 104 was previously activated.

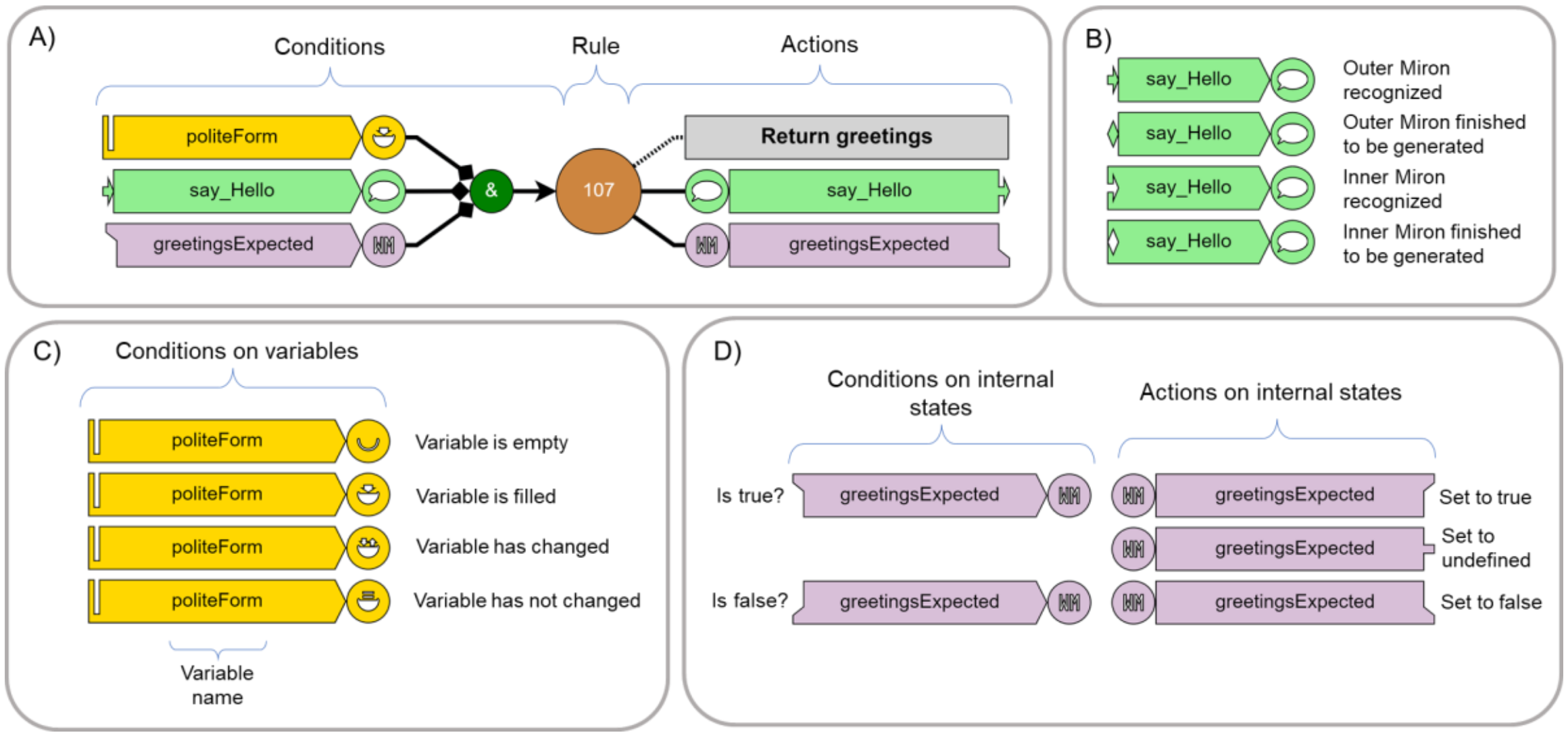

Figure 11: Rule DSL Elements : A) Graphical representation of a rule defined by its autogenerated id (107), its conditions and its actions (see text) the gray box is just a comment used to explain the rule. B) Possible state of a Miron that can be used as part of a condition. C) Possible state of a variable that can be used as part of a condition. The change detection of variable content is based on the comparison of previous and current time step. D) Possible conditions and actions on internal states. Internal states are represented by neural levels 1 for true, -1 for false and no activity for undefined. Only the levels 1 and -1 can be tested since undefined is not represented in the system.

An exception to the previous conditions is applied whenthere are incoming inhibitory conditions (red links). In the example, if the internal state “inhibit_Speech” is true, this condition alone will inhibit the rule with higher priority to all other non-inhibitory conditions.

Looking at the action part of rule 125, the rule triggers two actions, it sets the internal state “greetingsExpected” to false AND generate the Miron “say_Hello” (to chat or TTS). However, since the rule has connections to other rules (145, 147), once activated it will be kept in the rule memory for follow up steps until one of the fanout rules gets activated, in which case the memory state of rule 125 will be deactivated. In the case of the example, both rules 145 and 147 have the same condition namely rule 125 is active in memory AND the Miron “say_Hello” is finished to be generated. In this case, when the condition becomes active, either rule 145 or rule 147 will be randomly activated as described in §3.1.

## 5.2. Code Generation

Once the model is created, the code generation phase translates the model into JavaScript files, where the rule file encodes the weights and connectivity of the model in a representation used in the behavior engine network (see right part of Figure 12). As shown in the figure, connectivity between rules is translated into a connectivity using the rule memory. Note that this representation does not show all the computational steps described in §3.1. Action connectivity for a given rule is translated into a parallel activation of Mirons or internal states actions. Complex logical conditions are translated into the activations of multiple AND nodes connected to a single OR node at the rule. Inhibitory links are represented by negative weights.

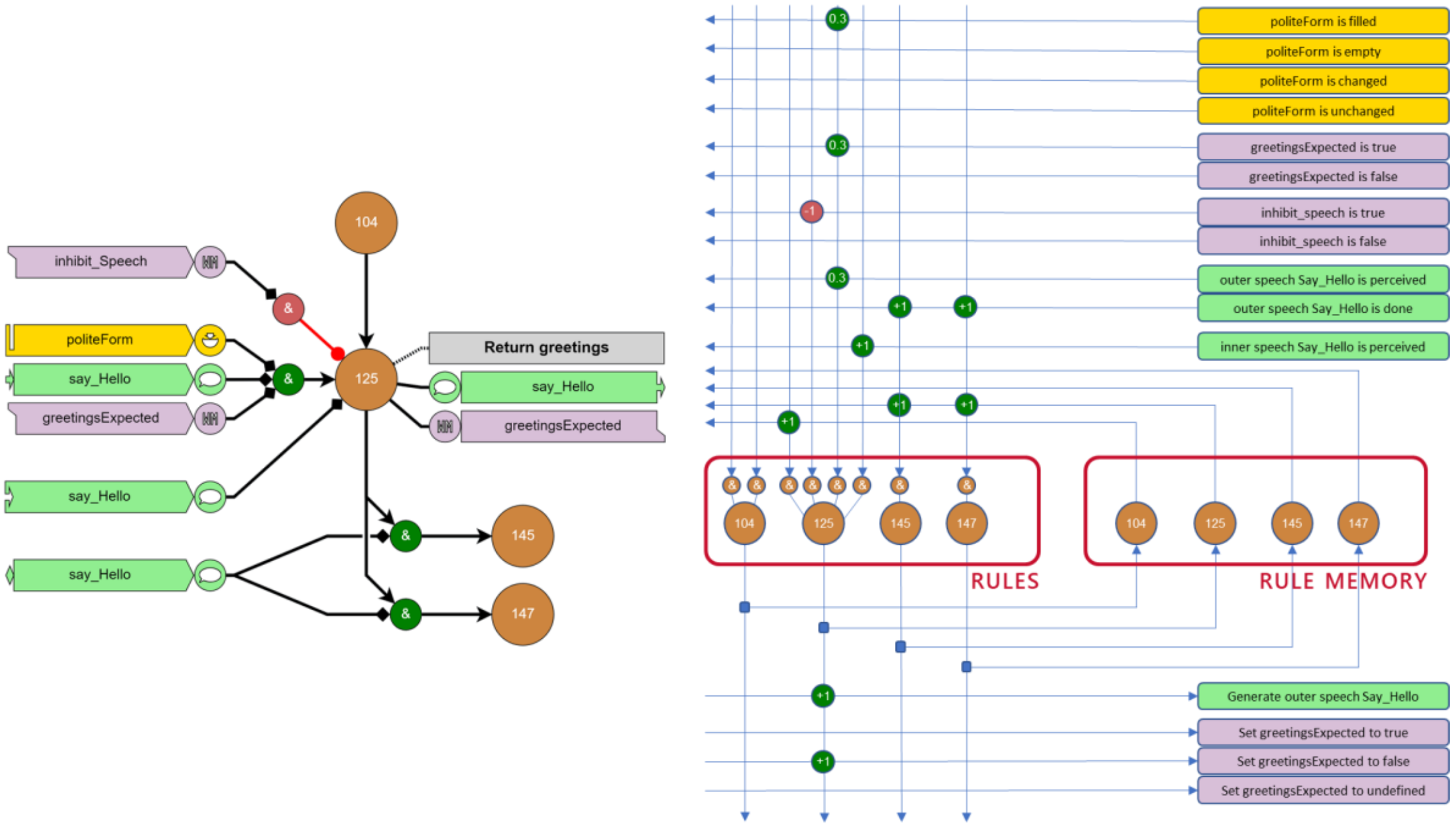

Figure 12: Example of rule nodes connectivity (left), and corresponding neural network connectivity in the behavior engine

The code generation process, also translates grammar model of Mirons into the Miron definition file, used by the Miron system to perform recognition and generation of text.

In the next section, we discuss the typology of architectures that can be designed to efficiently encode dialog and behavior applications.

## 6. Dialog Design

Dialog in specialized domains, can be seen as a succession of speech acts (Morelli et al. 1991)

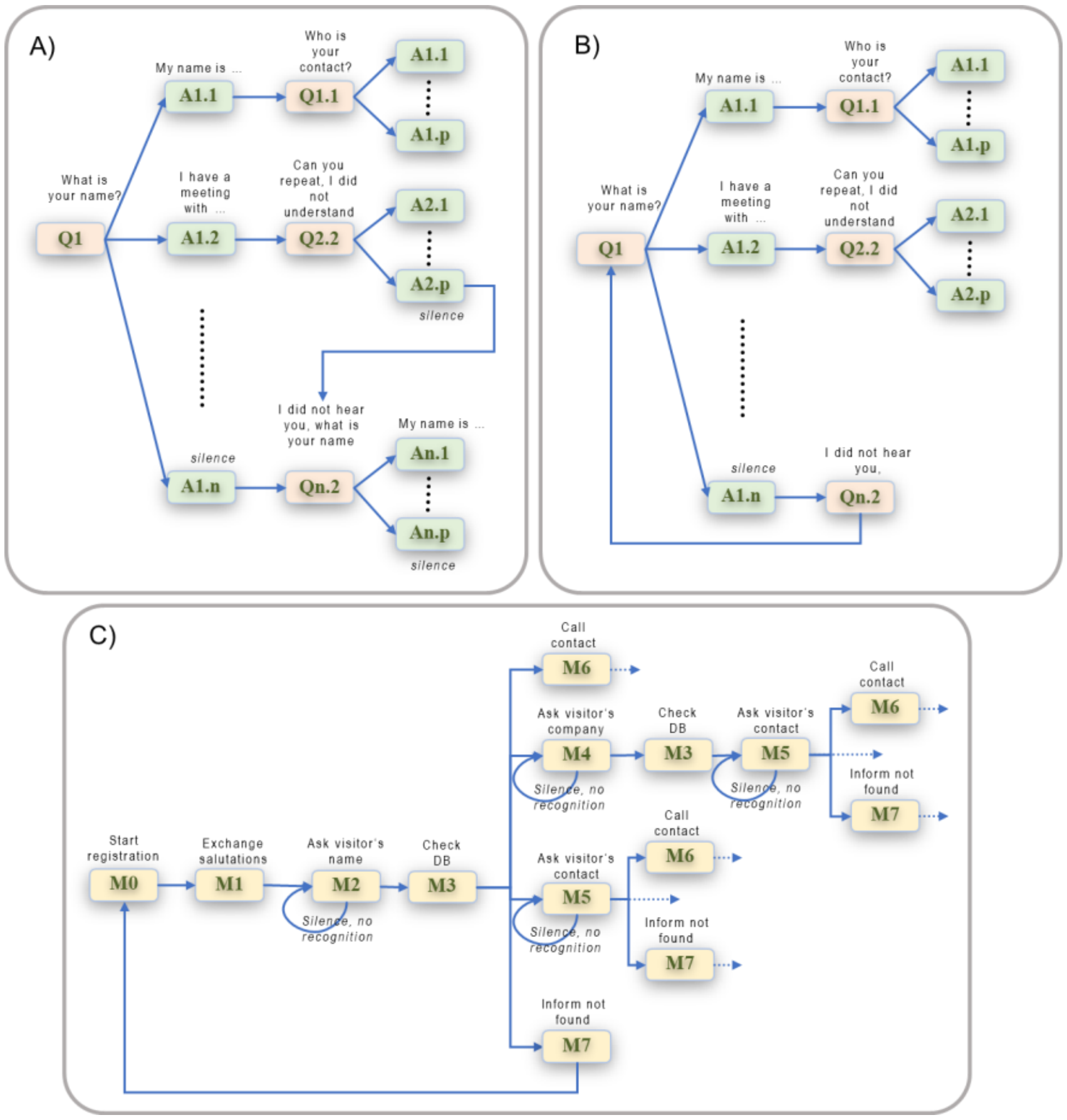

Figure 13: Basic dialog architecture. The examples given here relates to a receptionist scenario but could be applicable for other types of application domain A) Acyclic graph of questions/answers. B) Cyclic graph of questions/answers. C) Modular architecture with reuse of specialized modules

between the machine and the user. When starting from an initial null-context, this succession of user/machine utterance can be seen as an acyclic graph where after each user utterance the system decide in which branch of the graph the dialog can continue (Figure 13, A). Branching can depend on user input but also on dialog context. Such a representation of dialog is of course very inefficient due to the combinatoric explosion that such graph can generate. In particular, similar state can reoccur at different place in the graph which is why a slightly better approach is to transform the representation to a cyclic graph where different answers can lead back to the same question for example (Figure 13, B). But such a representation, although more compact still suffers combinatoric explosion. The next move in dialog design consist in identifying recurrent, similar dialog patterns, which is a well-known process in programming, namely modularization (Figure 13, c). Indeed, the same sub-dialog can be initiated from different places in the dialog graph and this re-usability of part contribute to simplify the dialog structure by a process of hierarchization and functional decomposition. Although important, this step still keep the dialog graph complex and huge.

### 6.1. Turn-taking loop

An important characteristic of dialog is the alternation between system and user interventions. Turn-taking covers different aspects of human-human and human-machine interaction, because timing, decision making, or multi-modality plays an important role in it. Some dialog systems like Gandalf, a planetary presentation system have focus explicitly on those aspects (Thorisson, 2002).

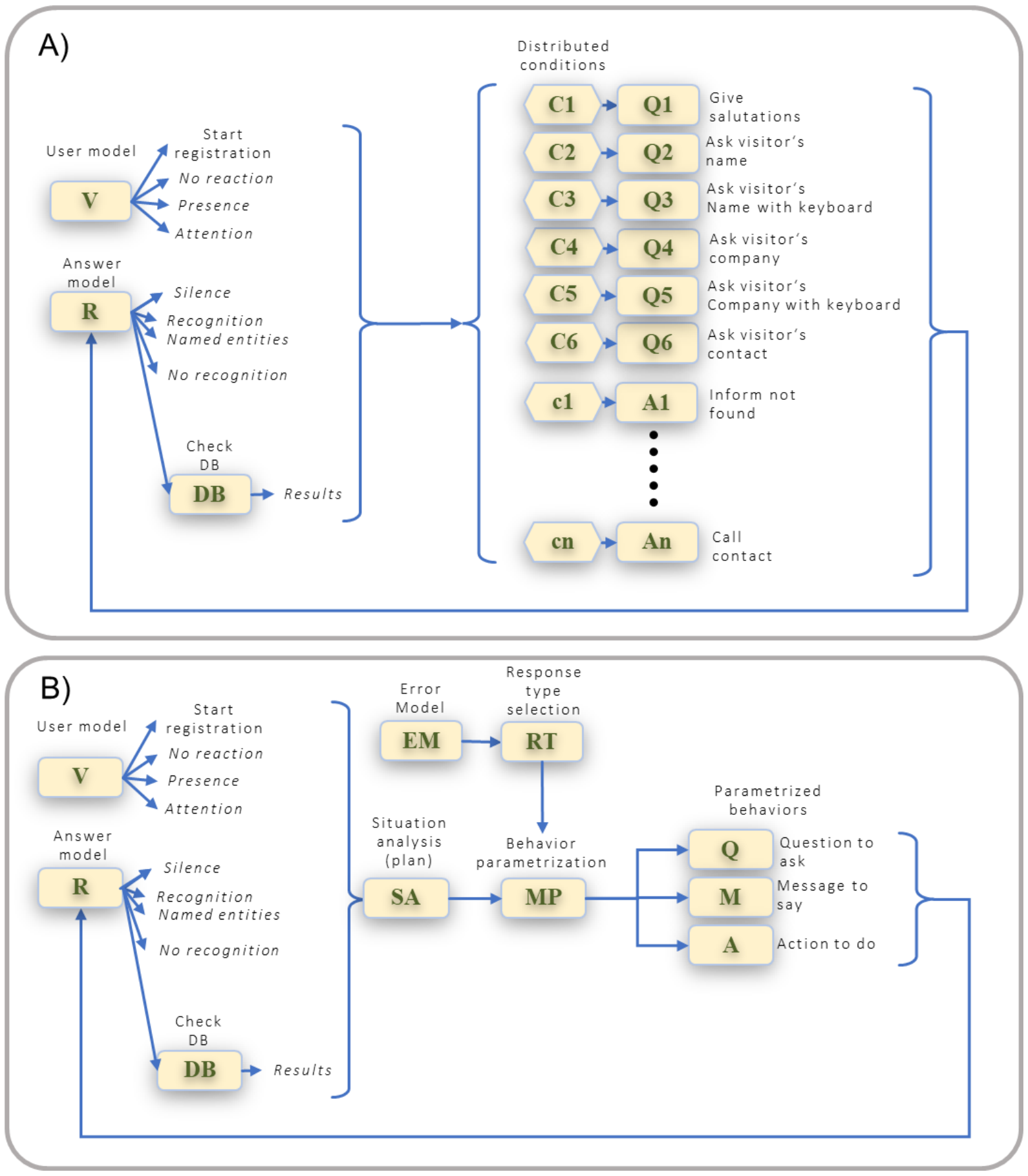

Figure 14: Turn-taking based dialog organization. Again, the example given here is drawn from a receptionist scenario. A) The reasoning part is represented by an undifferentiated pool of actions associated to some conditions. B) Use of modularization, and parametrization to represent generic structures of the dialog flow.

From a dialog design point of view, turn-taking can be exploited to simplify the design of dialog management architecture by creating a main loop that represent each turn of the dialog (Figure 14). Such an architecture resembles the perception-reason-action loops found in other domain of AI and robotics (Cutsuridis & Taylor, 2013). It reduces drastically the combinatoric of the dialog graph by folding it back into a recursive structure.

In a first step the reasoning part can simply be represented by a pool of all actions or speech acts needed for the application associated to their condition of activation (Figure 14, A). This absence of structure does not scale and makes the design extremely error prone to mis-specified conditions that often lead to unwanted spurious parallelization of actions. Therefore, other principles must be introduced to get a grip on dialogs complexity.

### 6.2. Planning, modularization & parametrization

Here again, neuroscience is instructive on how behavior combinatoric can be modularized. Action representation in the brain is decomposed in several functionalities that go from planning a task in the pre-frontal part of the brain (Tanji & Hoshi, 2001) to preparing or specifying the corresponding actions in pre-motor cortices (Hoshi & Tanji, 2007) and triggering the action in the motor cortex. This decomposition translated into dialog and behavior design means separating a planning phase representing abstract sequence of needed functions from a preparation phase defining the parameters of the actions and from the representation of a limited number of parametrized actions (e.g. by intent, or named entity variables).

By decomposing the problem in this way (Figure 14, B), it is possible to define generic modules independent of the task or domain and nevertheless fundamental for the integrity of the system and improving its designability. Such modules can be for example:

- A "user model", responsible for the analysis of the user response, his presence and attention can be acquired through non-auditory channels like camera and other sensors, the quality of the auditory environment (in case of ASR usage in noisy places) or simply reaction time measurement (in chat for example)
- A "response model" used to check degree of recognition in the current context,
- An "error model" dealing with misunderstandings or absence of user reaction to define redundant modality to use (voice, keyboard, buttons) and appropriate reaction to use (re-asking, giving help, …) in a similar way as error handling is done in (Bohus & Rudnicky, 2009) for example.
- A "situation analysis" module playing the role of a plan specifying "what" should be the next action or the next question answer independently of the "how" which should be realized in an "behavior parametrization" module. Such a separation of concern is also used in system like the Ravenclaw DM (Bohus & Rudnicky, 2009) or (Wu et al., 2001)
- A "behavior parametrization" module specifying "how" a behavior should be performed and defining in working memory the semantic of the parameters used to execute the behaviors (Miron to use, query…).
- A set of generic "parametrized behavior" like providing information, asking a question, proposing alternative input modality (speech, text field and keyboard, buttons, gesture…) but also making DB queries, placing a phone call, or sending emails. These behaviors are generic in the sense that the semantic content is defined in the parameter of these functions

These principles can be used to guide the development of domain specific applications, reducing complexity for a designer but still nevertheless insufficient to allow scaling to free dialogs and therefore requiring further research. A similar kind of task decomposition has been used in (Nakano et al., 2005) for example, to cope with the interplay between dialog management and robot behavior planning demonstrating its applicability.

We now turn to a concrete application implemented following the previously mentioned principles and techniques.

# 7. Application

To test the above presented approach, we have implemented a virtual receptionist used at the entrance of our research institute. The system is running on a booth and is embodied by a virtual avatar supposed to great visitors, get information about them, verify that they are planned to visit the institute and call a contact person to fetch them at the entrance. This section describes shortly the architecture of the system, some aspects of the behavior and dialog design and ends with a user study qualifying the user experience with the system.

## 7.1. A Virtual Receptionist

Virtual receptionists are an interesting type of application used by several authors either as information kiosk, to study interaction patterns and how to engage conversations with visitors wanting to order shuttle buses (Bohus & Horvitz, 2009), to study how greetings influence the type of further communication with a virtual receptionist (Lee, Kiesler, & Forlizzi., 2010) or multi-modal interactions between machine and humans (Babu et al, 2006). Most of such system use embodiments to interact with humans, either in form of animated virtual avatar or in form of complex robots (Sakagami et al., 2002). Here, we design a virtual avatar displayed on a touch screen and chose to focus on the behavior and dialog interaction.

### 7.1.1. System Architecture

The system was designed to be easy to develop for, having most of the logic in the code running directly in the browser. To add interactions with third party systems and to accommodate for security restrictions in modern browsers, a node.js server was used (see Figure 15).

Node.js was chosen to support more easily migrating components from client side to server side when the need would arise (both server and client use Javascript).

Below we describe the main components of the system.

**The avatar framework:**

The avatar framework represents the pluggable system that allows the user to work with the different components of the system as well as the parts needed to create new applications. It contains adapters for external systems to allow easy interaction from browser code as well as the avatar themselves and scene editing capabilities: adding objects, moving, animating etc. The main entry point for all client interactions with the system is a component called System Loader which aims to abstract the overall pluggable architecture of the framework and provide the user a single unified interface.

**Speech technologies:**

The avatar framework intergrates both TTS and ASR. The TTS part is handled by a proprietary system provided by Acapela™ ( https://www.acapela-group.com ); the user presented with a simple Javascript interface which allows for generating utterances in all the languages supported by the Acapela server. The speech recognition part is handled by a thin Javascript adapter that talks directly to a Speechmatics™ ( https://www.speechmatics.com/ ) server via websocket. The speechmatics server also offers multiple languages as well as the ability to add your own words to the engine (e.g., visitor and contact persons' proper names, as well as company names in this application). The system works fully offline, as it does not rely on any cloud services.

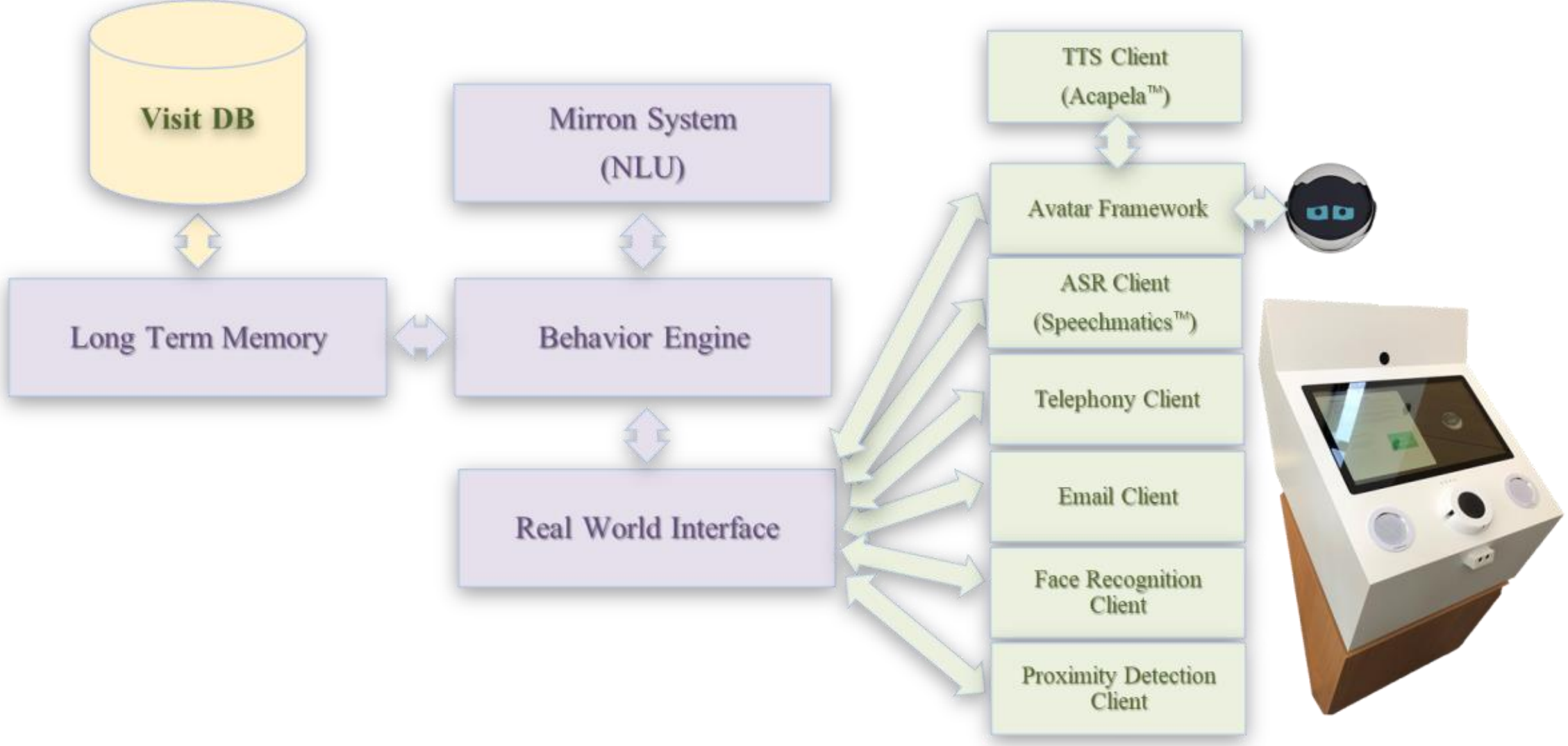

Figure 15: System architecture running an avatar receptionist on the booth depicted on the right.

**Communication via phone or email:**

Additionally, there also is two ways for the avatars to communicate with the external world: via phone and via email. An avatar can place a phone call and use TTS and speech recognition to have a "conversation" with the person on the other end. If the phone system fails for any reason, it fallbacks to sending emails via a simple email sending component. The system also supports receiving emails, although not used in this application.

**Facial recognition and proximity detection**

As part of offering as an intuitive user interface as possible, the avatar framework also supports facial recognition to be used to "remember" a user who interacts with the system. Of course, this is optional, and the developer of an application based on the framework must implement GDPR compliant techniques. No images of the user are stored locally, only feature vectors for the recognized face is stored so that the AI system can recognize a user when they come back.

Moreover, the system has support for an Arduino based proximity sensor so that camera can be turned on or off the camera for life-time management of the component and for GDPR regulation, i.e. only activating camera when needed.

#### 7.1.2. DIALOG ERGONOMICS

Following the concepts described in section §6, we designed the dialog in a modular way (see Figure 16). The dialog is based on mixed initiative and organized in a big turn taking loop where the system first identifies the state of the visitor's presence, response, attention, and face recognition state (visitor model), then analyses the situation (situation analysis) based on slot content and database query results and formulate messages (message parametrization) to both inform about its state of knowledge and request missing information. The error handling part allows four steps of recovery: In step one the planned question or information is given, in case of error (e.g., misunderstanding, absence of response…) the system switch to step 2 where the context is explained, and the question reformulated. If error persist, the system proposes to change modality (e.g., switching from voice to keyboard or buttons). By persisting errors, the system proposes a last chance to answer via a virtual keyboard or buttons depending on the situation (e.g., name misunderstood can be typed, multiple alternatives can be chosen via buttons on the screen) after which the system eventually closes the dialog, sending the visitor to the human receptionist. When the dialog is successful (i.e., the visitor is found in the list of registered visitors) the system calls the contact person of the visitor or the institute secretariat or send emails in case of unsuccessful calls. Then the visitor is asked to read and accept or deny the usage of his biometric data for research purpose. In case the visitor accepts the disclaimer, his/her facial features are saved and can be used to recognize him/her on a next visit and shorten accordingly the registration dialog.

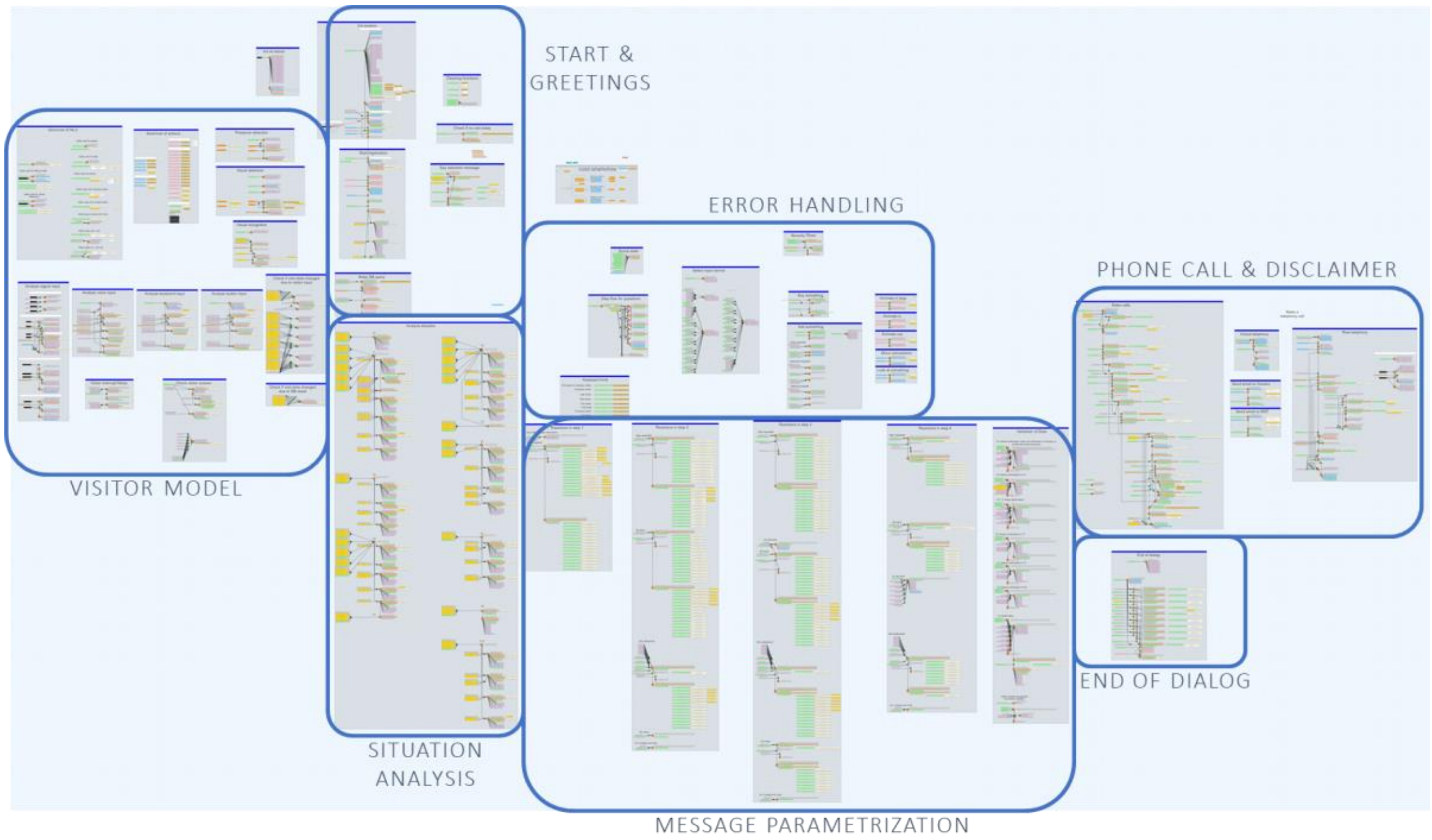

Figure 16: Graph of the virtual receptionist dialog. Only the main parts of the graph are made visible in the picture to illustrate the structure.

Redundancy is at the heart of the design as explained in the way error are handled, but also by providing sub-titles for the avatar utterances.

On the behavior side, besides animated actions like placing calls or sending emails, the avatar express facial emotions in situations where it could make sense (e.g., by unsuccess or success in registration, in calling contacts or acceptance or deny of disclaimer), it also moves towards or away from the visitor depending on dialog states and visitor proximity. These behaviors follow the concept developed by (Funakoshi at al., 2010) stating that an embodied dialog system needs to provide a quick reaction to user utterances, but this reaction does not necessarily need to be verbal to improve user impression and reduce speech collisions.

#### 7.1.3. Testing using Simulation

From a software development aspect, the creation of simple simulators for the different functional modules (e.g., long-term memory, visual recognition, telephony, emails…) reveals to be an important way to test system internal APIs and allow parallel development by limiting module dependencies.

In a similar way as the interlocutor simulation described by Lee, Yung et al. in 2010, we implemented a simulation mode in which the same dialog DSL could be used to generate visitor behavior and emulated dialogs. This functionality allows to create test cases and test suites for the interactions.

## 7.2. User Study

To evaluate the user acceptance of the virtual receptionist application and the quality of the dialog with the virtual receptionist, a user study was conducted using some standard in usability measurements. Evaluation was performed using the System Usability Scale (SUS) (Brooke, 2013), the AttrakDiff scale (Hassenzahl et al., 2003) and a specific scale.

### 7.2.1. Participants

Due to the strong reduction of visitors at our institute in 2020 due to covid 19 regulation in Europe, 20 participants were recruited among company associates to play the role of visitors. The participants joined voluntarily and none of them received any compensation. Furthermore, the participants had no a-priori knowledge of the system and were not taught how to use it. 40% of the participants were female, 60% male, aged in mean 34,5 ±7,4 years old; 75% were German with one French, one Italian, one Russian, and two Chinese, all fluent in English. To avoid participants to meet each other or to watch each other doing the test, a gap of 30 minutes was set between two test sessions.

### 7.2.2. Test Task Scenario

After picking randomly two pre-defined visitor profiles specifying visitor names, company names and contact person, the participants had to try to register two times with the system using the chosen profiles. A successful registration was terminated with the virtual receptionist calling the contact person of the visitor. This phase was done fully alone using English language. After performing the task, the participants were asked to fill out alone the three questionnaires (SUS, AttrakDiff and the specific one). Questionnaires were available both in English and German language. All questionnaires were anonymous.

### 7.2.3. Test Environment

The booth running the virtual receptionist is located in the entrance hall of our building, near the desk of a human receptionist. The hall is a big circular room 12m in diameter and 6m in height, covered with reflecting surface producing quite an amount of acoustic echo.

The noise environment of the entrance was not controlled, and background noise was possible (side conversations between company personal, automatic sliding door opening and closing, noisy roll of trolley on tiled floor across the room…)

### 7.2.4. Specific Scale

Beside using standard evaluation scale like SUS and AttraktDiff, a specific scale was design to get a user feedback more dedicated to the specificity of the virtual receptionist application. Considering some insight from psychology to the design of user interface (Weinschenk, 2011) a questionnaire of 30 questions (see Figure 18) was build based on the following categories and using a 7-point Likert scale (Joshi et al., 2015):

- People don’t want to work or think more than they must (question 1)
- People have limitations (questions 2 & 3)

- People make mistakes (questions 4 to 6)
- Memorizing is difficult (questions 7 & 8)
- People are social (questions 9 to 11)
- People are easily distracted (questions 12 & 13)
- People crave information (questions 14 to 20)
- People's brains engage in unconscious processing (question 21)
- People create mental models (questions 22 to 24)
- People cannot find information if presented in a cluttered way (questions 25 to 28)
- People are judgmental (question 29 & 30)

The questionnaire was created in English and then translated in German, so that participants can have a form as close as possible to their native language.

### 7.2.5. Test Results

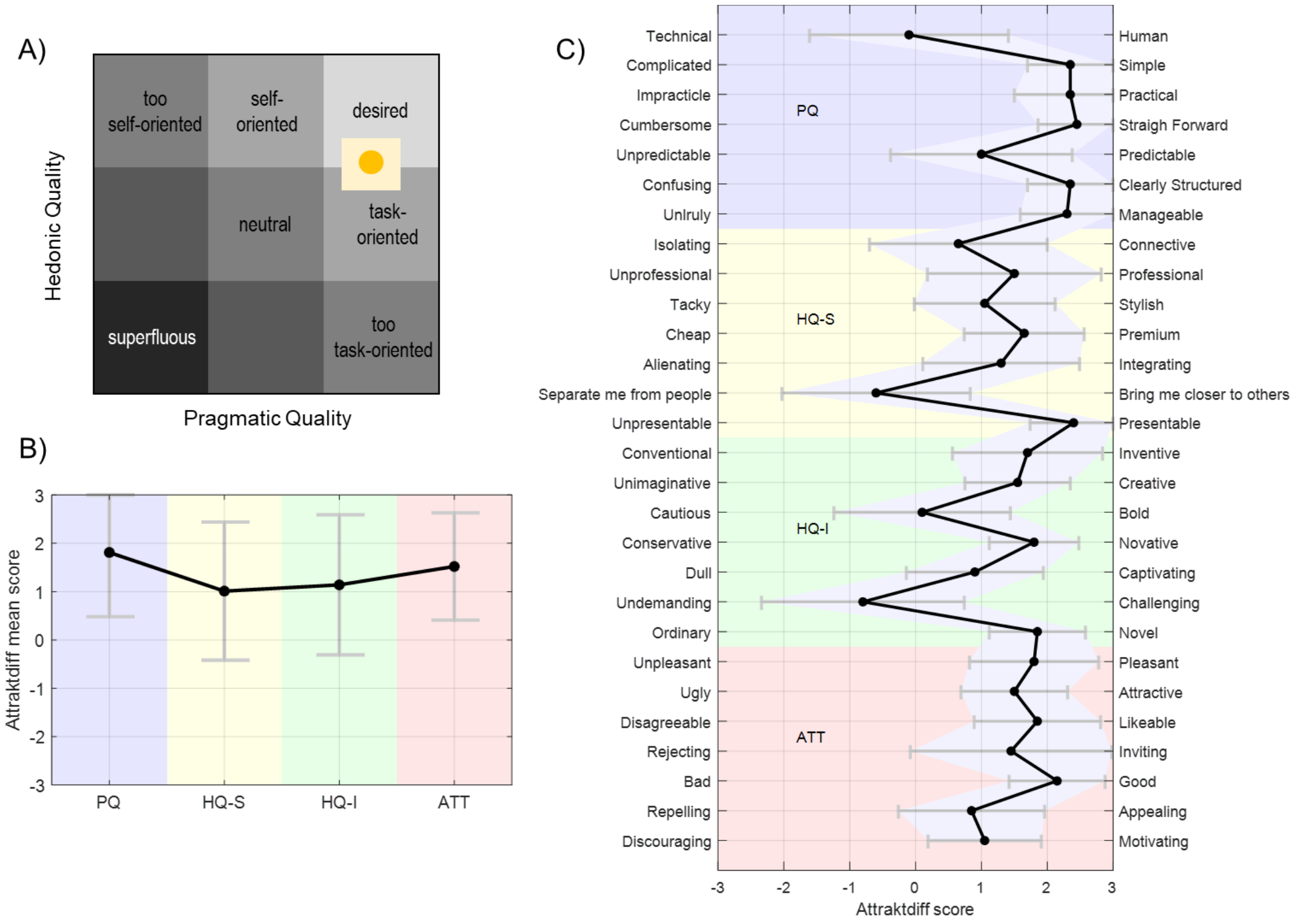

Figure 17: Result of the AttraktDiff scale. A) Portfolio-presentation with average value and confidence surface. B) Mean scores with confidence interval. C) Word pair detail result.

All participants managed to register properly. The SUS scale gave a score of 90,3 ±10,3 which sets the system in the highest range of the acceptance scale. The AttrakDiff facilitates a more precise evaluation of products by customers or users. The evaluation data enables to gauge how the attractiveness of the product is experienced, in terms of usability and appearance and whether optimization is necessary. The AttrakDiff scores are computed over 4 qualities:

- Pragmatic Quality (PQ): Is the system easy to use? Does it help the user to reach the goal?
- Hedonic Quality – Identity (HQ-I): Does the system have a good image for the user?
- Hedonic Quality – Stimulation (HQ-S): Is the system stimulating the user?

- Attractiveness (ATT): Is the system attractive (combining pragmatic and hedonic qualities)?

AttrakDiff scores can be represented using three diagrams (see Figure 17):

- The portfolio (defining the "desirability" of the system along the Pragmatic and Hedonic axes),
- The average scores of each quality
- Word pairs (providing detail results on different aspects of the system).

Overall, the results are in the positive range of the scale with superior values for the pragmatic qualities of the system. The detailed results show generally positive words associations with a quite high variance. Exceptions concern the undecidedness on the perception of the system as "technical/human" and "undemanding/challenging" which we find positive for a technical system. The notion of "separate from people" versus "bring closer to people" is understandable for a system that somehow replace a function traditionally performed by humans.

Finally, the specific scale is presented in Figure 18. Overall, the positive results of SUS and AttraktDiff are confirmed in this scale (questions 9, 21, 22, 27, 29, 30). Specific questions on the quality of the speech recognition, speech synthesis (questions 2,3,21, 25,26) and dialog comprehension and support (questions 4, 5, 6, 10, 11, 13, 22, 28) were also very positively scored even with a low variance. The set of questions 14 to 20 and 23 and 24 goes beyond the focus of this paper and are related to one function of the virtual receptionist which is to provide information on usage of privacy data by the system since it uses biometric data like voice, and facial images.

## 8. Conclusions & Further Work

In this paper, we have tried to show that even hand-crafted dialog management system can benefit from some concepts inspired from biology like, mirror neurons, recurrent architecture based on multiple feedback loops, inner speech, or multi-modality. Designing dialog requires tools and methods to avoid the natural combinatorial explosion of language and we showed that model driven development approaches as well as graphical DSL could help but should be combined with particular care in the design principles used to reduce this natural complexity. We are well aware that only learning methods will allow for scalable solutions in the future and even though current learning methods shift the problem to the amount of training data, they are the direction to go for sure. We believe that dialogs are the product of human interactions and that therefore efficient learning methods will need to be inspired by the research in infant development and that imitation learning, instructional learning or curriculum learning will play an important role there. At least our research will turn in this direction.

### Acknowledgements

Many thanks go to colleagues of Rinftech, Andrei Hutuca, Cristian Sandu, Andrei Jifcovici, Alexandru Stefanoiu, Alecsandru Patrascu and Iulia Oprea for their contribution in the realization of the Avatar framework, to Christophe Lorin for his contribution to the database, dialog design and user study, to Nicholas Moderson and Roberto Fichera for their help on the telephony part, and to the colleagues of HRI-EU that contributed to various aspects of the project: Siddhata Naik, Mark Dunn, Oliver Schön, Stefan Fuchs, Nils Einecke, Manuel Dietrich and the IT department of HRI-EU. This research was financed by Honda Research Institute Europe.

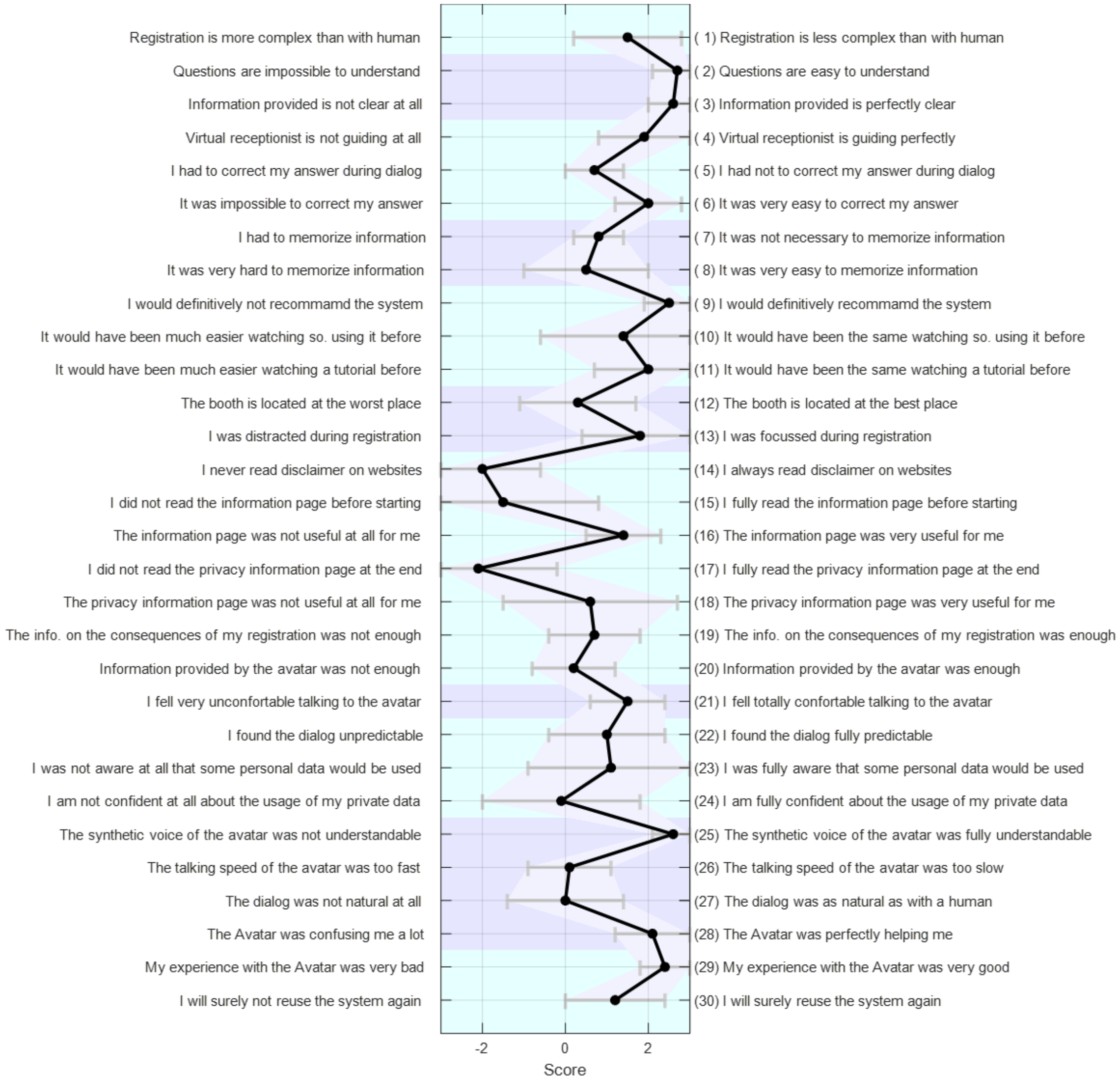

Figure 18: Result of the specific scale.